\documentclass[conference]{IEEEtran}
\IEEEoverridecommandlockouts
\usepackage{cite}
\usepackage{amsmath,amssymb,amsfonts}
\usepackage{algorithmic}
\usepackage{graphicx}
\usepackage{textcomp}
\usepackage{xcolor}
\usepackage{hyperref}
\usepackage{color, soul}
\usepackage{balance}

\usepackage[labelformat=simple]{subcaption}

\DeclareCaptionLabelSeparator{periodspace}{.\quad}
\usepackage{listings}
\def\BibTeX{{\rm B\kern-.05em{\sc i\kern-.025em b}\kern-.08em
    T\kern-.1667em\lower.7ex\hbox{E}\kern-.125emX}}

\begin{document}

\title{LAMS: LLM-Driven Automatic Mode Switching for Assistive Teleoperation
\thanks{*Both authors contributed equally to this research.

This material is based upon work supported by NIST under Grant No. 70NANB23H216 and by the National Science Foundation under Grant No. 2341352.}
}

\author{\IEEEauthorblockN{Yiran Tao*}
\IEEEauthorblockA{
\textit{Carnegie Mellon University}\\
Pittsburgh, PA, USA \\
yirantao@andrew.cmu.edu}
\and
\IEEEauthorblockN{Jehan Yang*}
\IEEEauthorblockA{
\textit{Carnegie Mellon University}\\
Pittsburgh, PA, USA \\
jehan@cmu.edu}
\and
\IEEEauthorblockN{Dan Ding}
\IEEEauthorblockA{
\textit{University of Pittsburgh}\\
Pittsburgh, PA, USA \\
dad5@pitt.edu}
\and
\IEEEauthorblockN{Zackory Erickson}
\IEEEauthorblockA{
\textit{Carnegie Mellon University}\\
Pittsburgh, PA, USA \\
zackory@cmu.edu}
}

\maketitle

\begin{abstract}
Teleoperating high degrees-of-freedom (DoF) robotic manipulators via low-DoF controllers like joysticks often requires frequent switching between control modes, where each mode maps controller movements to specific robot actions. Manually performing this frequent switching can make teleoperation cumbersome and inefficient. On the other hand, existing automatic mode-switching solutions, such as heuristic-based or learning-based methods, are often task-specific and lack generalizability. In this paper, we introduce LLM-Driven Automatic Mode Switching (LAMS), a novel approach that leverages Large Language Models (LLMs) to automatically switch control modes based on task context. Unlike existing methods, LAMS requires no prior task demonstrations and incrementally improves by integrating user-generated mode-switching examples. We validate LAMS through an ablation study and a user study with 10 participants on complex, long-horizon tasks, demonstrating that LAMS effectively reduces manual mode switches, is preferred over alternative methods, and improves performance over time. The project website with supplementary materials is at \url{https://lams-assistance.github.io/}.
\end{abstract}


\section{Introduction}

One of the key challenges in robotic teleoperation systems is mapping a controller's limited degrees of freedom (DoF) to a robot’s higher DoF, especially for high-DoF robotic manipulators. This challenge is particularly prominent in assistive applications, where users with motor impairments rely on low-DoF assistive devices such as tongue-based joysticks~\cite{mohammadi2019high, mohammadi2021continuous}, head orientation systems~\cite{padmanabha2024independence, padmanabha2023hat}, eye gaze controls~\cite{alva2017image, chin2008integrated}, and sip-and-puff systems~\cite{mougharbel2013comparative, grewal2018sip} to perform daily tasks.

\begin{figure}
    \centering
    \includegraphics[width=\linewidth]{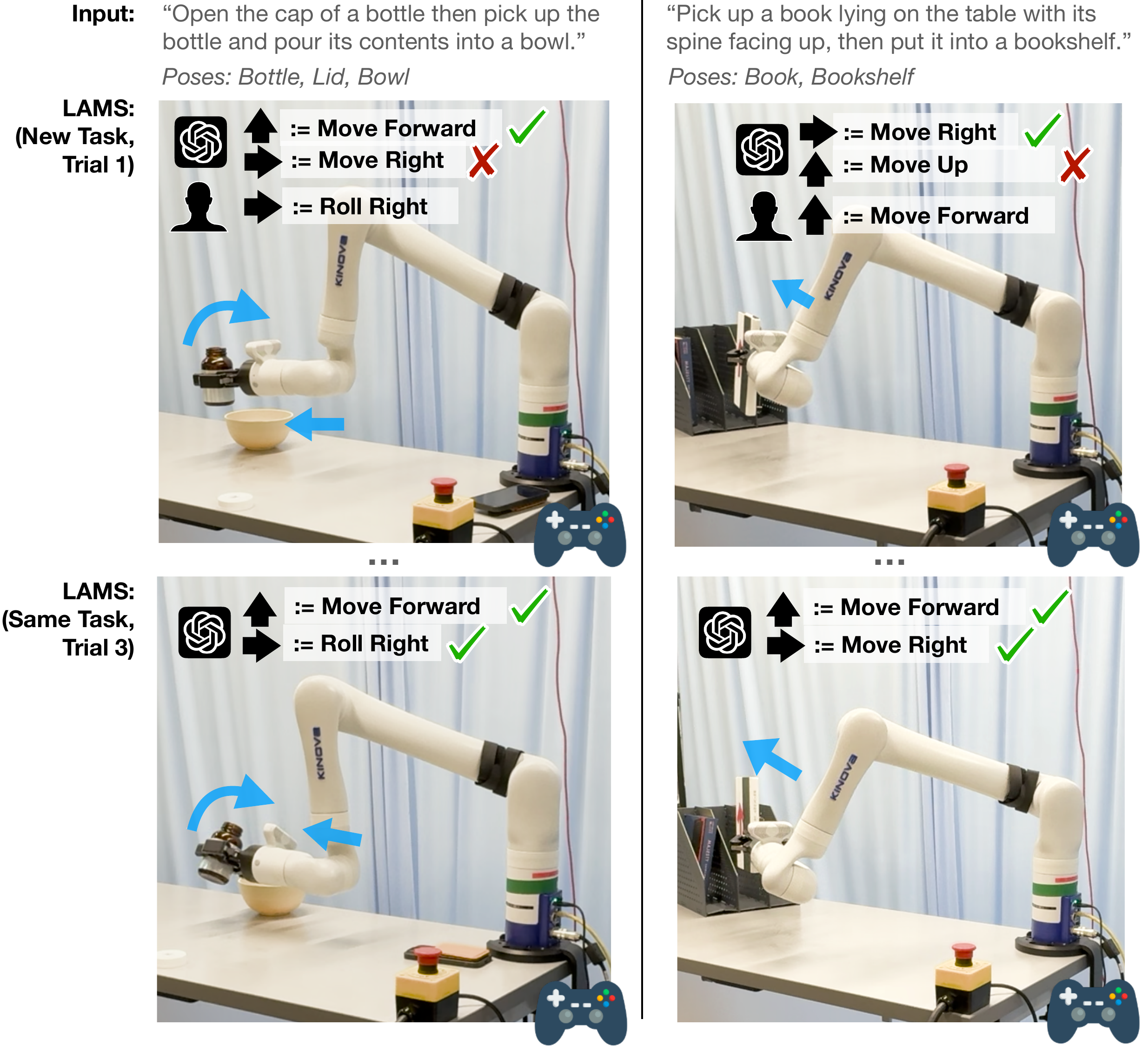}
    \caption{We introduce LLM-Driven Automatic Mode Switching (LAMS), which uses Large Language Models (LLMs) to automatically predict the most effective mapping between joystick and robot movement directions. LAMS requires no prior task demonstrations and incrementally improves as the user repeatedly interacts with the system. \textbf{Top}: In the initial trials, while able to provide useful mapping predictions, LAMS encounters some errors due to limited task knowledge, requiring users to occasionally perform manual mode switches. \textbf{Bottom}: By the third trials, with LLM prompts enhanced by integrating prior user manual switches, LAMS performs automatic mode switches accurately with minimal user intervention.}
    \label{fig:lams_introduction}
    \vspace{-1.5em}
\end{figure}

Simple controllers, such as 2-DoF joysticks, typically require users to switch between different control modes, where each mode defines a specific mapping of the joystick's four movement directions (up, down, left, right) to specific robot actions, such as translation, rotation, or gripper control.  This process becomes cumbersome when users must frequently switch modes to complete a long-horizon, multi-stage task, leading to inefficiency and cognitive strain. For instance, in teleoperating a robotic arm to place a book on a shelf (as shown in the right two subfigures in Fig.~\ref{fig:lams_introduction}), the user must switch among different modes for each subtask: translating and rotating the robot end effector to align with the book, closing the gripper to grasp the book, translating and rotating again to align the book with the shelf, and finally placing the book. These frequent mode switches interrupt the workflow, forcing the user to repeatedly recall and select the correct mode, which increases frustration, cognitive load, and reduces task efficiency~\cite{herlant2016assistive, yang2023high, arrington2004cost}.

To address this, automatic mode switching aims to handle these transitions seamlessly, allowing users to focus on the task itself rather than the control mechanism.
Prior works have explored automating mode switching in teleoperation and assistive robotics using heuristic-based approaches~\cite{gopinath2017mode, quere2020shared}, reinforcement learning~\cite{rl, kizilkaya2024intelligent}, and optimization techniques~\cite{herlant2016assistive}. However, many of these solutions are task-specific, requiring demonstrations or hand-engineered rules for each task, limiting their generalizability to new scenarios.

Large Language Models (LLMs), which have demonstrated strong performance on a wide range of applications \cite{llm1, llm2, llm3, llm4, llm5}, offer a promising solution to these challenges with their commonsense reasoning capabilities~\cite{commonsense}. In this work, we propose leveraging  LLMs for automatic mode switching, with the assumption that LLMs can utilize their rich contextual understanding ability to make effective mode-switching predictions in novel tasks, without requiring pre-collected demonstrations or hand-engineered rules.

\begin{figure*}[h!]
    \centering
    \includegraphics[width=\linewidth]{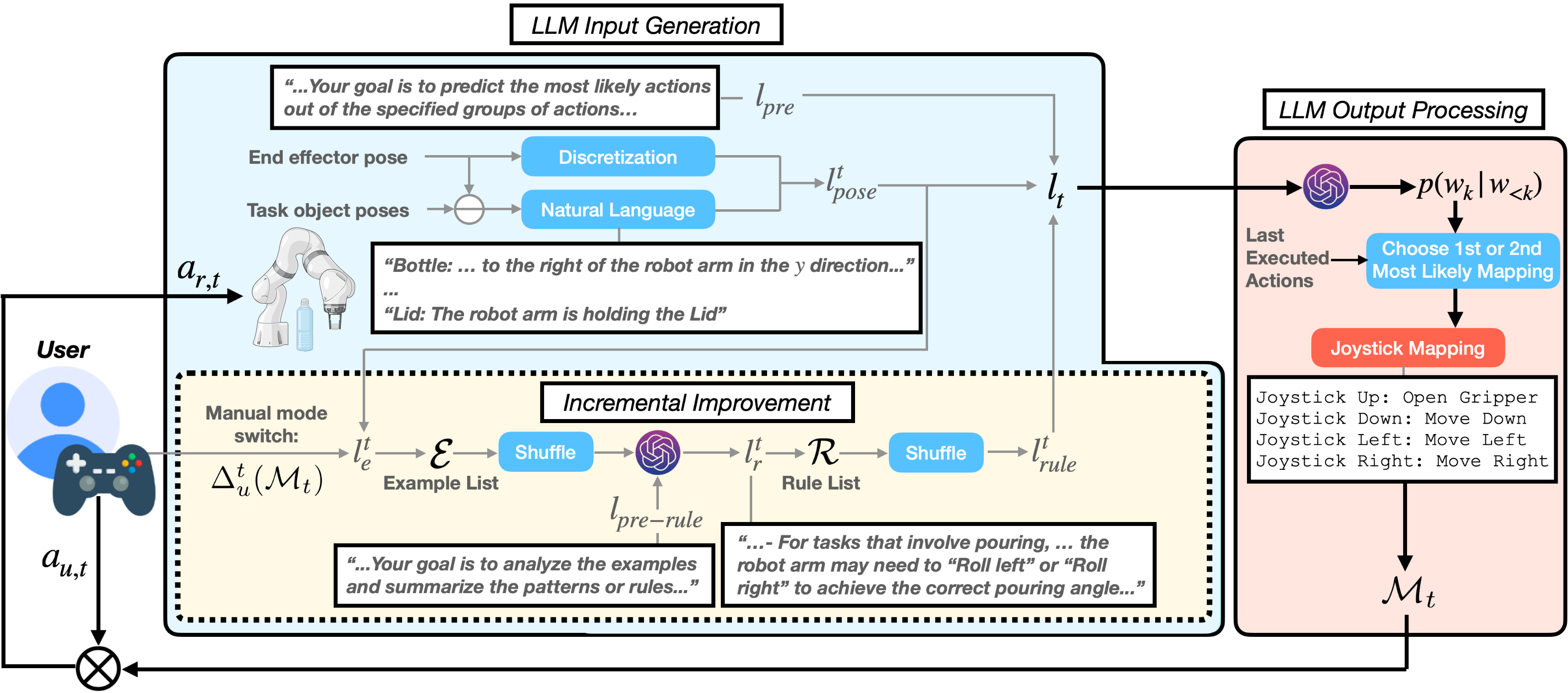}
    \caption{Our proposed LLM-Driven Automatic Mode Switching (LAMS) framework. LAMS grounds the current robot end effector and task object poses into a natural language description $l_{pose}^{t}$. This description, along with a prompt prefix $l_{pre}$ and a rule prompt $l_{rule}^{t}$, forms a natural language instruction $l_{t}$, which is fed into an LLM to generate the mode $\mathcal{M}_{t}$, i.e., the mapping of the joystick’s four movement directions to specific robot action directions. $\mathcal{M}_{t}$, along with user action $a_{u,t}$ produces robot action $a_{r,t}$. LAMS begins without task-specific demonstrations, and improves incrementally through user interaction by incorporating user-generated examples into the rule prompt $l_{rule}^{t}$. The framework consists of three main components: LLM Input Generation, LLM Output Processing, and Incremental Improvement, which are respectively detailed in Section~\ref{sec:input},~\ref{sec:output} and~\ref{sec:incre}.}
    \label{fig:flowchart}
    \vspace{-1.5em}
\end{figure*}

Building on these insights, we introduce LLM-Driven Automatic Mode Switching (LAMS), which eliminates the need for task-specific data or predefined heuristics. As illustrated in Fig. \ref{fig:flowchart}, LAMS translates the current task context into a natural language instruction, which is fed into an LLM to perform mode switching, i.e., predict an effective mapping of the joystick’s four movement directions to specific robot action directions. Additionally, LAMS improves incrementally as the user interacts with the system by incorporating user-generated mode-switching examples into its language instructions. We demonstrate the effectiveness of LAMS' design decisions through a quantitative ablation study.


To formally evaluate LAMS, we conducted a user study where 10 participants used a Kinova robotic arm to perform two complex, long-horizon tasks. Study results support our hypotheses that (1) LAMS enables users to complete complex multi-stage tasks with fewer manual mode switches, and is preferred over alternative mode-switching methods; and (2) LAMS improves its automatic mode-switching ability over time as a user repeatedly performs a task, in contrast to a static LLM-based method.

In summary, the key contributions of this paper are: \begin{enumerate} 
    \item We introduce LAMS, a novel LLM-driven framework for automatic mode switching that eliminates the need for task-specific demonstrations or predefined heuristics. 
    \item We design LAMS to incrementally improve through user interaction, by incorporating user-generated mode-switching examples into the LLM prompts. 
    \item We conducted extensive evaluations of LAMS, including ablation studies and a real-world user study, demonstrating that LAMS effectively reduces manual mode switches, is preferred by users over alternatives methods, and improves performance over time.
\end{enumerate}

\section{Related Works}

\subsection{Automatic Mode Switching}
Automatic mode switching has been explored in several prior works to reduce the cognitive load and inefficiencies associated with manual mode switching. For example, Herlant et al.~\cite{herlant2016assistive} proposed a time-optimal model that uses time as a cost metric, utilizing Dijkstra’s algorithm to predict when the robot should automatically change modes. However, as they pointed out in their own paper, algorithms like Dijkstra's, which compute the optimal cost-to-go, incur combinatorial computational costs relative to the size of the search space, making them impractical for high-dimensional systems like high-DoF robotic arms in real-world applications.

More recently, Gopinath et al.\cite{gopinath2017mode} proposed an approach that performs mode switching by placing the user in control modes that maximally disambiguate between various goals in the scene. However, this approach is dependent on effective human-intent recognition, which remains an ill-defined and challenging problem. Quere et al.\cite{quere2020shared} took a different approach, dividing tasks into multiple phases, each with different motion constraints and input mappings. In order to do so, this approach requires extensive hand-engineering to define the task phases, constraints, and mappings, limiting its scalability and flexibility across different tasks.

In addition, Pilarski et al.\cite{rl} and Kizilkaya et al.\cite{kizilkaya2024intelligent} leveraged reinforcement learning (RL) for automatic mode switching. These methods, however, require substantial training before the RL agent can be deployed, limiting their real-world applicability. Kizilkaya et al.~\cite{kizilkaya2024intelligent} additionally rely on a real-world dataset for training, which further restricts their generalizability to new environments or tasks.

Beyond these limitations, most existing automatic mode switching methods are task-specific, requiring tailored demonstrations or hand-engineered rules, which limits their scalability and generalizability to novel tasks. In contrast, our framework LAMS, leverages the commonsense reasoning capabilities of LLMs to eliminate the need for task-specific data or predefined heuristics, allowing it to generalize across tasks.

\subsection{Learned Latent Action Models}
Another related line of research addressing the challenge of mapping a controller’s limited DoF to a robot’s higher DoF utilizes learned latent action models~\cite{losey2022learning, jeon2020shared, karamcheti2022lila, latent0, latentwo}.

These approaches train auto-encoders to tackle the challenge: During training, the encoder compresses high-dimensional robot actions into a latent space matching the low-DoF controller, while the decoder reconstructs the original high-dimensional actions. During deployment, the user’s control inputs are fed into the decoder to generate the corresponding robot actions.

While these methods provide an alternative to mode switching in addressing the teleoperation challenge, they typically rely on extensive training datasets with substantial expert demonstrations and, in some cases, user-annotation processes to ensure intuitive control~\cite{losey2022learning}. This data collection can be costly, hard to scale, and challenging to adapt to new tasks and environments. Notably, one of these works attempted to minimize the need for human demonstrations~\cite{latentwo}, but their user study revealed that ``users were confused when the unsupervised robot learned unexpected behaviors''.

In contrast, our approach reduces the cost and complexity of data collection while maintaining flexibility and scalability across different teleoperation tasks. Moreover, it preserves user intuitiveness, making it a preferred solution over alternative mode switching methods, as demonstrated in our user study.

\subsection{LLMs for Robot Planning and Control}
Recent work has achieved great success in utilizing LLMs to generate both planning and control signals for robots, such as decomposing high-level task descriptions to mid-level plans~\cite{inner, socratic, song2023llm,huang2023grounded}, generating robot code~\cite{liang2023code, wu2023tidybot, prog, voyager}, producing sequences of end-effector poses~\cite{discretize}, and selecting motion primitives~\cite{control1}. LLMs have also been integrated into robot systems that interact with humans. For example, Mahadevan et al.~\cite{mahadevan2024generative} used LLMs to generate and modify expressive robot behaviors such as nodding to interact with humans. Padmanabha et al.~\cite{padmanabha2024voicepilot} integrated LLMs into a feeding robot. Pandya et al.~\cite{pandya2023chatgpt} introduced an LLM-enabled interface for surgeons to control robotic tools.

While these approaches represent substantial advances in using LLMs for robot planning and control, there is minimal work exploring the use of LLMs to facilitate human teleoperation of robots. Our work seeks to bridge this gap by leveraging LLMs for automatic mode switching, allowing users to teleoperate robots more effectively with reduced manual mode switches.

\section{LAMS: LLM-Driven Automatic Mode Switching}

\subsection{Problem Statement}
\label{sec:statement}
We consider the scenario where a human operator teleoperates a high-DoF robot arm to perform a manipulation task with a low-DoF controller. This task is modeled as a sequential decision-making process, defined by tuple $(\mathcal{S}, \mathcal{A}_{u}, \mathcal{A}_{r}, \mathcal{T})$, where $s_{t} \in \mathcal{S}$ denotes the task state at time step $t$, $a_{u,t} \in \mathcal{A}_{u} \subseteq \mathbb{R}^{m}$ denotes the user action, $a_{r,t} \in \mathcal{A}_{r} \subseteq \mathbb{R}^{n}$ denotes the robot action, with $m \ll n$. $\mathcal{T}$: $\mathcal{S} \times \mathcal{A}_{r} \rightarrow \mathcal{S}$ is an unobserved transition function. 


In our experiments, the robot's action space is 7-dimensional (i.e., $n=7$), represented by the vector:
\[
a_{r,t} = 
    (\Delta x_{t}, \Delta y_{t}, \Delta z_{t}, \Delta \text{roll}_{t}, \Delta \text{pitch}_{t}, \Delta \text{yaw}, \Delta \text{gripper}_{t})
\]
where $\Delta x_{t}$, $\Delta y_{t}$, $\Delta z_{t}$ are the deltas in the Cartesian coordinates, $\Delta \text{roll}_{t}$, $\Delta \text{pitch}_{t}$, $\Delta \text{yaw}_{t}$ are the deltas in the Euler angles, and $\Delta \text{gripper}_{t}$ is the delta in the gripper's opening.

We use a joystick with two degrees of freedom ($m=2$) as the human control interface. The user action $a_{u,t}$ represents joystick movements, which is a 2-dimensional vector:
\[
a_{u,t} = (x_{u,t}, y_{u,t})
\]
where $x_{u,t}$ represents the joystick's lateral movement, and $y_{u,t}$ represents the joystick's longitudinal movement.


Our goal is to define a function $\mathcal{F}(s_{t}, a_{u,t})$: $\mathcal{S} \times \mathcal{A}_{u} \rightarrow \mathcal{A}_{r}$ that transforms the task state and user input into a robot action that is optimal for the task. Since we are focusing on mode-switching, we aim to generate an effective mapping $\mathcal{M}$ that aligns each joystick movement direction with a corresponding robot action direction, as illustrated in Fig. \ref{fig:interface}. Specifically, in our setting, joystick movement directions map to the following robot action directions, depending on the current control mode:

\begin{figure}
    \centering
    \includegraphics[width=0.95\linewidth]{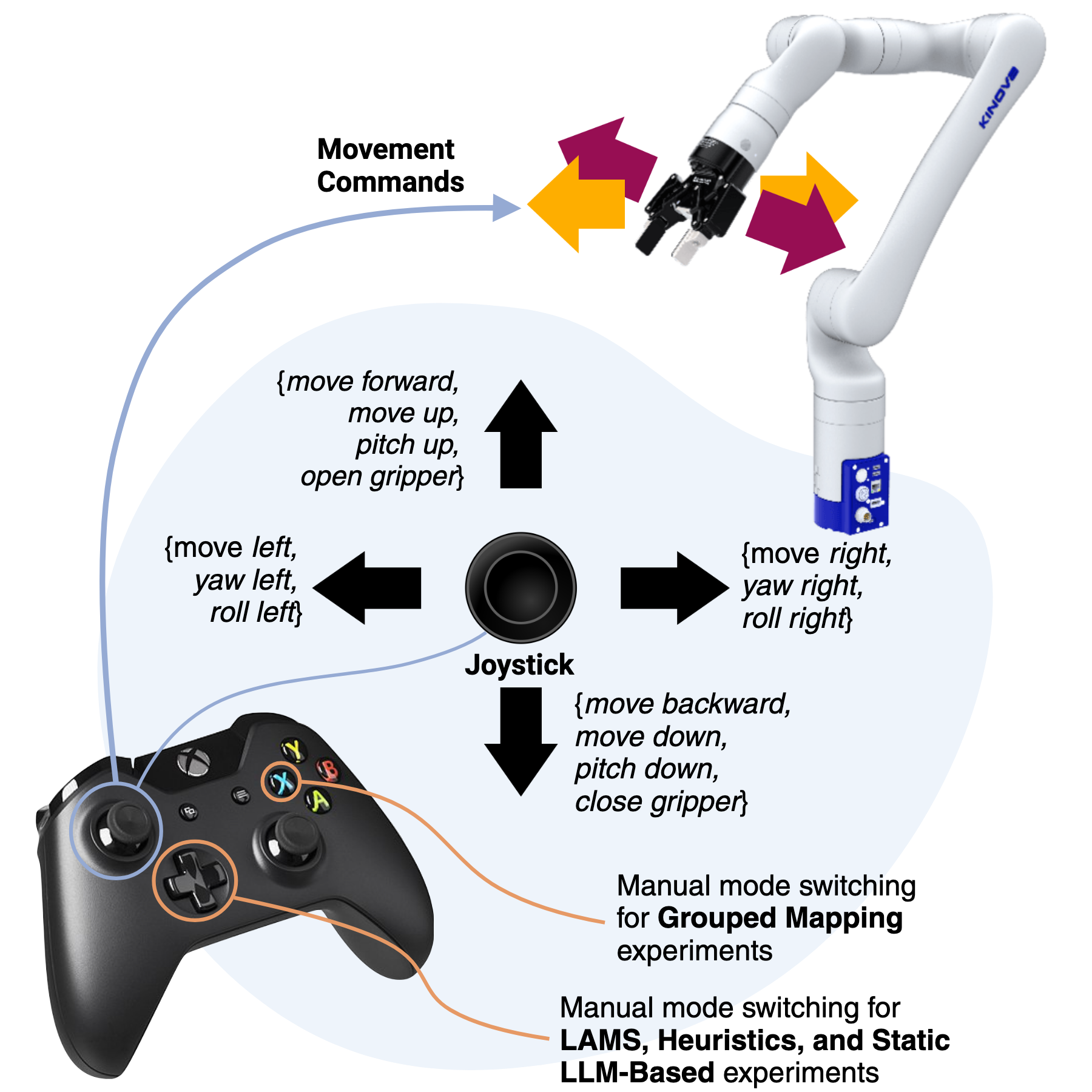}
    \caption{Usage of the Xbox controller as the user interface in our experiments.}
    \label{fig:interface}
    \vspace{-1.5em}
\end{figure}

\begin{itemize}
    \item \textbf{$\mathcal{D}_{up}$}: \{move forward, move up, pitch up, open gripper\},
    \item \textbf{$\mathcal{D}_{down}$}: \{move backward, move down, pitch down, close gripper\},
    \item \textbf{$\mathcal{D}_{left}$}: \{move left, yaw left, roll left\},
    \item \textbf{$\mathcal{D}_{right}$}: \{move right, yaw right, roll right\}.
\end{itemize}

Robot velocity is proportional to the magnitude of the user’s action $a_{u,t}$: Each element of $a_{u,t}$ is scaled by a constant scalar factor $v_m$ to control the velocity in the corresponding robot action direction based on the control mode. The index $m \in \{tr, ro, gr\}$ specifies whether the action dimension corresponds to translation, rotation, or gripper opening/closing.

In other words, depending on the current control mode, a user action such as $(x_{u,t}, 0)$ would map to one of the following robot actions: $A_{x}$ = \{$(\Delta x_{t} = v_{tr} \cdot x_{u,t}, 0, 0, 0, 0, 0, 0)$, $(0, 0, \Delta z_{t} = v_{tr} \cdot x_{u,t}, 0, 0, 0, 0)$, $(0, 0, 0, 0, \Delta \text{pitch}_{t} = v_{ro} \cdot x_{u,t}, 0, 0)$, $(0, 0, 0, 0, 0, 0, \Delta \text{gripper}_{t} = v_{gr} \cdot x_{u,t})\}$. Similarly, a user action $(0, y_{u,t})$ would map to $A_{y}$ = \{$(0, \Delta y_{t} = v_{tr} \cdot y_{u,t}, 0, 0, 0, 0, 0)$, $(0, 0, 0, \Delta \text{roll}_{t} = v_{ro} \cdot y_{u,t}, 0, 0, 0)$, $(0, 0, 0, 0, 0, \Delta \text{yaw}_{t} = v_{ro} \cdot y_{u,t}, 0)\}$, where $v_{tr}$, $v_{ro}$, and $v_{gr}$ control translation, rotation, and gripper opening/closing velocities, respectively. A combined joystick movement $(x_{u,t}, y_{u,t})$ will map to a combination of these robot actions.





In this work, we leverage LLMs to perform the function $\mathcal{F}(s_{t}, a_{u,t})$. Specifically, we decompose this function into three stages:
$$
\mathcal{F}(s_{t}, a_{u,t}) = \mathcal{O}(LLM(\mathcal{I}(s_{t})), a_{u,t}),
$$
where $\mathcal{I}(s_{t})$ represents the grounding of the task state into a language instruction $l_{t} \in \mathcal{L}$, which is then input into an LLM. $\mathcal{L}$ is defined as the set of natural languages. The output of the LLM is processed by $ \mathcal{O}(\cdot, a_{u,t})$ to produce the robot action $a_{r,t}$. LAMS begins without task-specific demonstrations, meaning there are no predefined input-output examples. Our objective is to incrementally refine this function as the user interacts with the system. The complete pipeline of LAMS is shown in Fig.~\ref{fig:flowchart}. In the following sections, we detail how we generate the input to LLM, process the LLM output to generate $a_{r,t}$, and leverage user interactions to improve the system over time. All LLMs used in this work are powered by GPT-4o.

\subsection{Generating LLM Inputs}
\label{sec:input}
At each time step $t$, when mode switching is required, the input to the LLM is a language instruction $l_{t} = [l_{pre}, l_{rule}^{t}, l_{pose}^{t}]$, structured as three main components. 

$l_{pre}$ is a prompt prefix that provides context such as objectives and output format for the LLM. The exact $l_{pre}$ we use can be found in Appendix \ref{sec:prompts}.1.

$l_{rule}^{t}$ contains rules that guide the LLM's mode-switching prediction. These rules are derived from user-generated mode-switching examples. $l_{rule}^{t}$ is initially empty, incrementally growing as the user interacts with LAMS. Examples of $l_{rule}$ can be found in Appendix \ref{sec:prompts}.2. We discuss the details of how $l_{rule}$ is augmented through user interactions in Section~\ref{sec:incre}.

The final component, $l_{pose}^{t}$, provides a description of the current pose of the robot arm and task-relevant objects. For the robot arm, we encode its pose as a dictionary containing the end-effector's Cartesian coordinates and Euler angles, along with the gripper status. For task objects, we describe the relative position of each object with respect to the robot arm's end-effector across six dimensions using natural language statements. We provide additional details on the construction of $l_{pose}^{t}$, along with an instance from our experiments, in Appendix  \ref{sec:prompts}.3. Our ablation studies (Section~\ref{sec:ablation}) demonstrate that the natural language grounding of object states is more effective than numeric representations.

\subsection{Processing LLM Outputs for Mode Switching}
\label{sec:output}
When a mode switch is required, which in our experiments occurs either at the beginning of the task or when the user pauses for 1.5 seconds to signal a mode switch need, the LLM is prompted to predict the most likely action direction from each of the action direction groups: $\mathcal{D}_{up}$, $\mathcal{D}_{down}$, $\mathcal{D}_{left}$, $\mathcal{D}_{right}$, resulting in four predicted action directions. 

A naive mode-switching approach would be to directly map the joystick's four movement directions to the LLM's natural language response. However, in certain phases of a task, more than one action may be equally desired by the user. The LLM's response under such circumstances may appear ineffective to the user if the user has already executed one action from $\mathcal{D}_i$ and is expecting the system to switch modes to a different action mapping.

To address this, we leverage an essential mechanism of LLMs: rather than directly generating a single response, LLMs produce a probability distribution over possible next words, denoted as $p(w_k|w_{<k})$, where $w_k$ is the word generated at the $k^{th}$ position in the response. Utilizing this probability distribution, we can assess the likelihood of each robot action in $\mathcal{D}_{i}$. Formally, for each robot action with natural language representation $d_{i,j}$ in group $\mathcal{D}_{i}$, where $j$ denotes the index of the action within $\mathcal{D}_{i}$, we compute $p(d_{i,j}|o_{i})$, where $o_{i}$ represents the LLM's response preceding $d_{i,j}$. 

For each group, if the robot action with the largest $p(d_{i,j}|o_{i})$, i.e., $d_{i}^{*} = \arg\max_{d_{i,j}} p(d_{i,j} | o_{i})$, has \textit{just} been executed before the current mode-switching call, we check whether the probability of the second most likely action, $d^{**}_{i} = \arg\max_{d_{i,j} \neq d^{*}_{i}} p(d_{i,j} | o_{i})$, exceeds a predefined threshold (empirically set at 0.2). If so, we use $d^{**}_{i}$; otherwise, we use $d^{*}_{i}$. The effectiveness of automatic mode switching over probability distributions is demonstrated in our ablation study (Section~\ref{sec:ablation}).

Through this process, we obtain $d_{up}$, $d_{down}$, $d_{left}$, $d_{right}$, corresponding to the joystick's four movement directions. The final robot action $a_{r,t}$ is a combination of the actions selected from $A_{x}$ and $A_{y}$ (defined in Section~\ref{sec:statement}), depending on $d_{up}$, $d_{down}$, $d_{left}$, $d_{right}$ and the user's action $a_{u,t}$.

\subsection{Incremental Improvement via User-Generated Examples}
\label{sec:incre}
While LAMS is able to perform useful automatic mode-switching for an unseen task even in the first interaction, it can encounter errors due to limited task knowledge. To address this, we design LAMS to incrementally improve as the user interacts with the system. This is achieved by incorporating user-generated mode-switching examples to augment the rule prompt $l_{rule}^{t}$ defined in Section~\ref{sec:input}. 

Particularly, during task execution, a Graphical User Interface (GUI) continuously displays the current mode $\mathcal{M}_{t}$, which shows the four robot action directions mapped to each joystick movement. If the user is dissatisfied with $\mathcal{M}_{t}$, they can manually switch to another mode $\mathcal{M}'_{t}$ (details on how manual switching is performed in our real-world experiments are provided in Section~\ref{sec:exp}). This manual mode switch can be expressed as $\mathcal{M}_{t}' = \Delta_{u}^{t}(\mathcal{M}_{t})$, which is then converted into natural language format $l_{\Delta_u^{t}(\mathcal{M}_t)}$. For example, an instance of $l_{\Delta_u^{t}(\mathcal{M}_t)}$ initiated by the user in our experiments was:
\begin{lstlisting}
    { "Joystick Up": "Pitch up"}
\end{lstlisting}
This means that the user switched the mapping of ``Joystick Up'' from a less preferred action to ``Pitch Up''.

To facilitate LAMS' incremental improvement, we maintain an example list $\mathcal{E}$ for each task. Each time a $\Delta_u^{t}(\mathcal{M})$ is made, $l_{pose}^{t}$ (defined in Section~\ref{sec:input}) and $l_{\Delta_u^{t}(\mathcal{M})}$ form an example $l^{t}_{e}$ in natural language format, which is added to $\mathcal{E}$ (see Appendix \ref{sec:prompts}.4 for an instance of $l_e^{t}$).


Instead of directly incorporating these examples into the LLM’s inputs, our ablation studies (Section~\ref{sec:ablation}) show that summarizing them into mode-switching guiding rules leads to more effective and robust improvements for LAMS. To achieve this, besides the example list $\mathcal{E}$, we also maintain a rule list $\mathcal{R}$ for each task, which starts empty and grows over time as the user performs the task. With $\mathcal{E}$ and $\mathcal{R}$, every time a new example is added to $\mathcal{E}$, all examples are shuffled and fed to a separate LLM (distinct from the one used for automatic mode switching) along with a prompt prefix $l_{pre-rule}$, to generate rules that guide future mode-switching predictions (the exact $l_{pre-rule}$ we use can be found in Appendix \ref{sec:prompts}.5). This LLM autonomously generates a variable number of rules $l_{r}^{t} = \{l^{t}_{r,k}\}$, $k={1,2,...,N}$, where $N$ is the number of generated rules. For instance, one of the rules generated in our experiments was (see Appendix \ref{sec:prompts}.2 for more rules generated in our experiments):

\begin{lstlisting}
- For tasks that involve pouring, such as pouring contents from a bottle into a bowl, the robot arm may need to "Roll left" or "Roll right" to achieve the correct pouring angle, especially if the bottle is already being held.
\end{lstlisting}
$l_{r}^{t}$ are appended to $\mathcal{R}$.

We note that this rule list update process happens after each manual mode switch, not just at the end of a task. This allows LAMS to improve dynamically during task execution, even on a user’s first attempt. As we will discuss in Section~\ref{sec:discussion}, this design enables LAMS to outperform a static LLM-based method even on the first time the user interacts with the system.

With the rule list $\mathcal{R}$, when a mode switch is required, the rules in $\mathcal{R}$ are shuffled to form $l_{rule}^{t}$, which is used as part of the input prompt for the LLM as discussed in Section~\ref{sec:input}. In our current approach, rules are reset between tasks to ensure independent evaluation. However, we believe certain rules are transferable across manipulation tasks, such as reaching for and aligning with an object for grasping. Future work could explore a shared, adaptable ruleset leveraging cumulative user interactions across tasks.




\section{Experiments}
\label{sec:exp}
\subsection{Tasks and Experiment Settings}
\label{sec:setting}
All experiments were conducted using a Kinova Gen3 robotic arm, with users controlling the robot via an Xbox controller, as illustrated in Fig.~\ref{fig:interface}. Specifically, the left joystick on the controller is used to generate user actions $a_{u,t}$. The mapping between joystick movement directions and robot action directions is either determined by an automatic mode-switching method (e.g. LAMS) or manually adjusted by the user when they are dissatisfied with the automatic switch.

As shown in Fig.~\ref{fig:interface}, to perform a manual mode switch $\Delta^{t}_u(\mathcal{M})$ in LAMS, the user presses one of the directional buttons on the Xbox controller's D-pad. This action updates the mapping for the corresponding joystick direction without affecting the others. For example, pressing ``up" on the D-pad changes the mapping for pushing up on the joystick, while leaving the mappings for other directions unchanged. This manual mode-switching process enables the user to correct errors made by automatic mode switching methods to ensure task completion, while also providing examples that help LAMS improve incrementally over time. 

In both the ablation study and the user study, the primary evaluation metric is the number of manual mode switches made by the user, with fewer switches indicating more effective automatic mode switching.

We evaluated LAMS on two complex, multi-stage tasks:
\subsubsection{Water Pouring} Open the cap of a bottle, then pick up the bottle, and pour its contents into a bowl.
\subsubsection{Book Storage} Pick up a book lying on the table with its spine facing up, then put it into a bookshelf.

For each task, to assess LAMS's ability to incrementally improve, a complete experiment consists of three interactions, where the user completes the task three times with different object layouts each time. We refer to these interactions as ``trial 1'', ``trial 2'', and ``trial 3''.

\subsection{Ablation Study}
\label{sec:ablation}
Prior to conducting the user study, we performed an ablation study on the water pouring task to evaluate key design choices in LAMS. The following alternative designs were tested to assess their impact on system performance:
\begin{itemize}
    \item \textbf{Num-State}: Using relative numeric values to represent the spatial relationship between objects and robot end effector in $l^{t}_{pose}$, rather than natural language descriptions.
    \item \textbf{Top-Action}: Always switching to the most likely action predicted in the LLM's natural language response, in contrast to mode switching over probability distributions as discussed in Section~\ref{sec:output}. 
    \item \textbf{Direct-Examples}: Directly using user-generated mode-switching examples ($\mathcal{E}$) in the LLM input $l_{t}$, instead of rules ($\mathcal{R}$) generated from these examples. 
\end{itemize}

For the ablation study, one researcher ran five experiments for each model on the water pouring task. Each experiment consists of three trials with different task layouts. 

\begin{figure}[htbp]
\centering
\includegraphics[width=0.84\columnwidth]{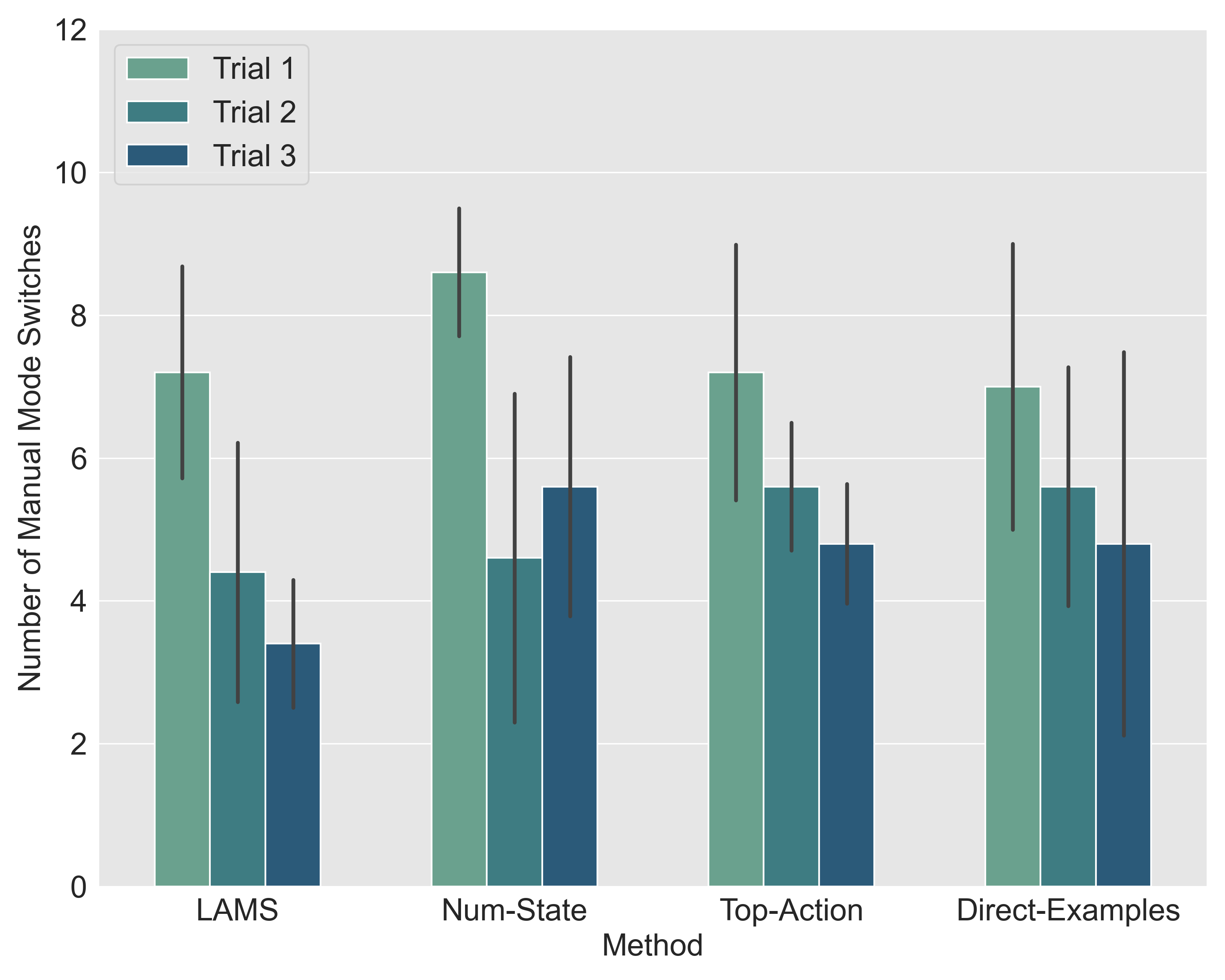} 
\caption{Average number of manual mode switches across 5 experiments from our ablation study on the water pouring task. Error bars show standard deviations.}
\label{fig:ablation}
\vspace{-1.5em}
\end{figure}

The average number of manual mode switches of different methods are shown in Fig.~\ref{fig:ablation}.

As shown in Fig.~\ref{fig:ablation}, LAMS steadily reduces the number of manual mode switches across trials. After just a few user interactions, the average number of manual mode switches drops from 7.2 in trial 1 to 3.4 in trial 3, a reduction of 52.8\%. In trials 2 and 3, LAMS consistently achieves the lowest average number of manual mode switches among all methods, with a low standard deviation, demonstrating both effectiveness and repeatability.

In contrast, the alternative methods show smaller reductions in the number of manual switches in trials 2 and 3. Notably, for Num-State, the average number of manual switches is higher than other methods on all three trials, indicating that it is easier for LLMs to reason from natural language descriptions than from numeric representations.

For Direct-Examples, trial 3 shows a high number of manual switches with a large standard deviation. This suggests as the number of user-generated examples grows, without summarizing them into rules, LLMs struggle to effectively determine which examples to reference for the current task state.

For Top-Action, trial 3 requires 1.4 more manual mode switches on average than LAMS, highlighting the benefit of LAMS' automatic mode switching over probability distributions, which considers both the most likely and second most likely actions to improve decision-making.


To further validate the effectiveness of these key design choices in LAMS, we perform a ``shadow mode'' analysis based on data collected from our user study. We provide details of this analysis in Appendix \ref{sec:shadow}.

\subsection{User Study}
\label{sec:user}

We conducted a user study with 10 participants (8 males, 2 females) aged 21 to 25 (mean age: 23.7), under a university-approved human subjects safety protocol. Participants reported an average experience level of 2 out of 7 with robotic arms and teleoperation, and 3.5 out of 7 with game controllers, where 1 represents no experience and 7 represents extensive experience. We test the following two hypotheses:
\begin{itemize} 
  \item \textbf{H1:} LAMS enables users to complete complex multi-stage tasks with fewer manual mode switches, and is preferred over alternative mode-switching methods.
  \item \textbf{H2:} LAMS improves its automatic mode-switching ability over time as a user repeatedly performs a task, in contrast to a static LLM-based method. 
\end{itemize}

\begin{figure*}[ht]
    \centering
    \begin{subfigure}[b]{0.42\textwidth} 
        \includegraphics[width=\linewidth]{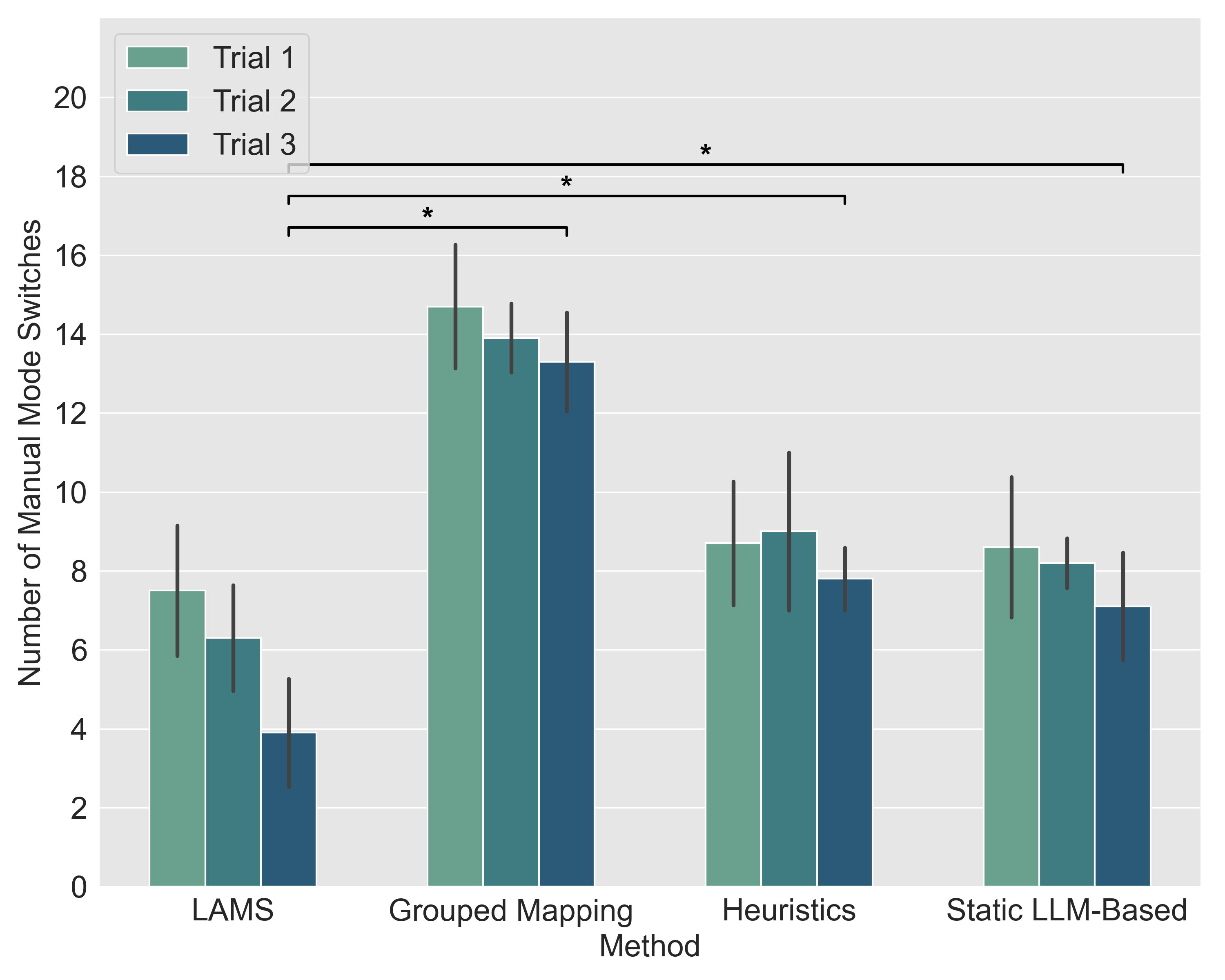}
        \caption{Water Pouring task}
        \label{fig:mode-switches-jar}
    \end{subfigure}
    \hspace{0.05\textwidth}
    \begin{subfigure}[b]{0.42\textwidth} 
        \includegraphics[width=\linewidth]{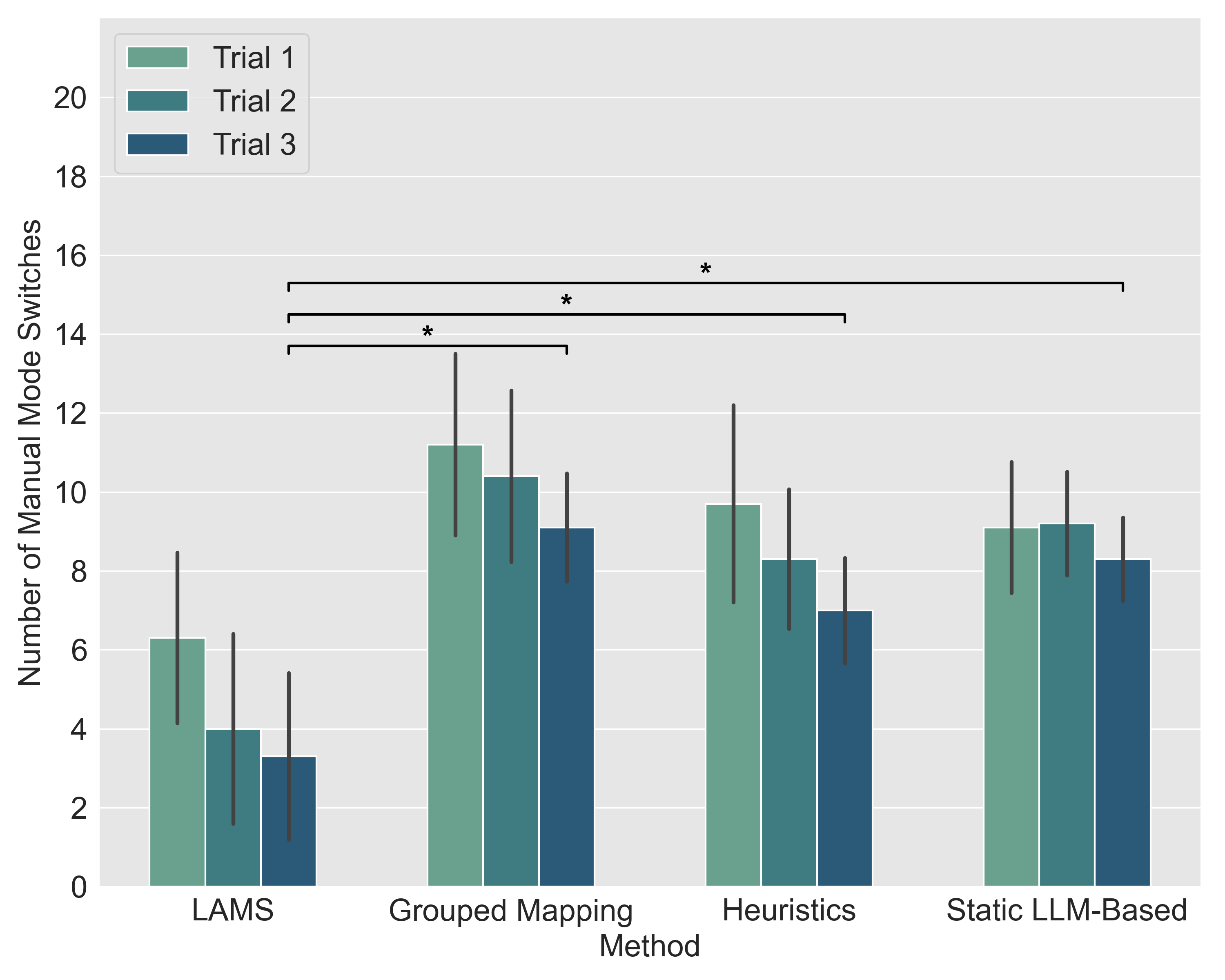}
        \caption{Book Storage task}
        \label{fig:mode-switches-book}
    \end{subfigure}
    \caption{Number of manual mode switches averaged over all participants. Error bars show standard deviations. Significance brackets indicate that there are statistical significant differences between LAMS and all other methods on trial 3 in both tasks.}
    \label{fig:mode-switches}
    \vspace{-0.5em}
\end{figure*}

To support these hypotheses, we compared LAMS with three baseline methods:
\begin{itemize}
    \item \textbf{Grouped Mapping}: This is a common mode-switching method in robotic arm applications \cite{harmonic, losey2022learning, latent0, yang2023high, padmanabha2023hat, padmanabha2024independence, gopinath2017mode, herlant2016assistive}, where robot actions are divided into predefined groups. Additional information on this method is provided in Appendix \ref{sec:grouped}

    \item \textbf{Hand-Engineered Heuristic Switching}: Inspired by~\cite{quere2020shared}, in this method, each task is pre-analyzed and manually divided into distinct subtasks, with optimal joystick mappings assigned for each subtask. Switching between subtasks is triggered by the robot's kinematic states, such as opening or closing of the gripper. When the task transitions to a new subtask, the pre-assigned mappings automatically switch accordingly.
    \item \textbf{Static LLM-Based Mode Switching (No Incremental Improvements)}: This method follows the LAMS pipeline but excludes $l^{t}_{rule}$ from $l_{t}$, meaning it remains static and does not improve through user interaction over time. This method is tested primarily to demonstrate that LAMS' improved performance is not merely due to users' increasing familiarity with the system, so as to provide evidence towards H2.
\end{itemize}

In our user study, for each task layout (i.e., each trial), users completed the task with all four methods in a counterbalanced order before moving on to the next task layout. The GUI is consistent across all methods. Appendix \ref{sec:gui} provides additional details and images of the experimental setup and the GUI.

We analyze the number of manual mode switches for each method, and user preferences for these methods in the following sections. Note that we did not analyze objective performance metrics such as task completion time or total end-effector travel distance, as in our study these were primarily determined by external factors like participants’ familiarity with the task and their level of attention, which were outside our study’s focus.

\subsubsection{\textbf{Number of Manual Mode Switches}}
The number of manual mode switches for each method averaged over all participants is shown in Fig.~\ref{fig:mode-switches}. 

Comparing LAMS with Grouped Mapping and Heuristic Switching, LAMS resulted in fewer average number of manual switches across all three trials for both tasks. Particularly in trial 3 of the water-pouring task, LAMS required an average of 3.9 manual switches, a reduction of 70.7\% compared to Grouped Mapping and 50.0\% compared to Heuristic Switching. Similarly, in the book storage task, LAMS required only 3.3 manual switches on average in trial 3, which is 63.7\% lower than Grouped Mapping and 52.9\% lower than Heuristic Switching. A Wilcoxon signed-rank test, adjusted using the Holm method, showed a statistically significant difference between LAMS and both Grouped Mapping and Heuristic Switching in trial 3 (corrected $p < 0.05$ for both tasks), and no significant difference (corrected $p > 0.05$) between the Static LLM and Heuristic methods. These results support the claim in H1 that LAMS enables users to complete complex multi-stage tasks with fewer manual mode switches compared to alternative mode-switching methods.

To test H2, as shown in Fig.~\ref{fig:mode-switches}, LAMS steadily reduced the number of manual mode switches across trials. After only 2 trials of water pouring, LAMS decreased the average number of manual mode switches from 7.5 to 3.9, a 48.0\% reduction. In the book storage task, the average number of manual mode switches dropped from 6.3 to 3.3, a 47.6\% reduction. To ensure that this reduction was not simply due to users' increasing familiarity with the system, we compare LAMS with the static LLM-based method: As shown in Fig.~\ref{fig:mode-switches}, LAMS consistently required fewer manual mode switches in both tasks. Particularly in trial 3, LAMS required 45.1\% less manual mode switches in the water pouring task and 60.2\% less in the book storage task compared to the static LLM-based method. A Wilcoxon signed-rank test showed a statistically significant difference between LAMS and the static LLM-based method in trial 3 ($p < 0.05$ for the water pouring task and $p < 0.01$ for the book storage task), providing evidence towards H2. Additionally, a generalized linear mixed model analysis on mode switches across tasks, conditions (LAMS vs. static LLM), and trials revealed a significant condition-trial interaction (coefficient = -1.075, p = 0.003, 95\% CI: [-1.789, -0.361]). This indicates a steeper reduction in manual mode switches over time for LAMS compared to static LLM, supporting H2. Notably, as shown in Fig.~\ref{fig:mode-switches}, LAMS outperforms the static LLM-based method even in the first trial. We discuss this result in Section~\ref{sec:discussion}.



\begin{figure}[htbp]
\centering
\includegraphics[width=\columnwidth]{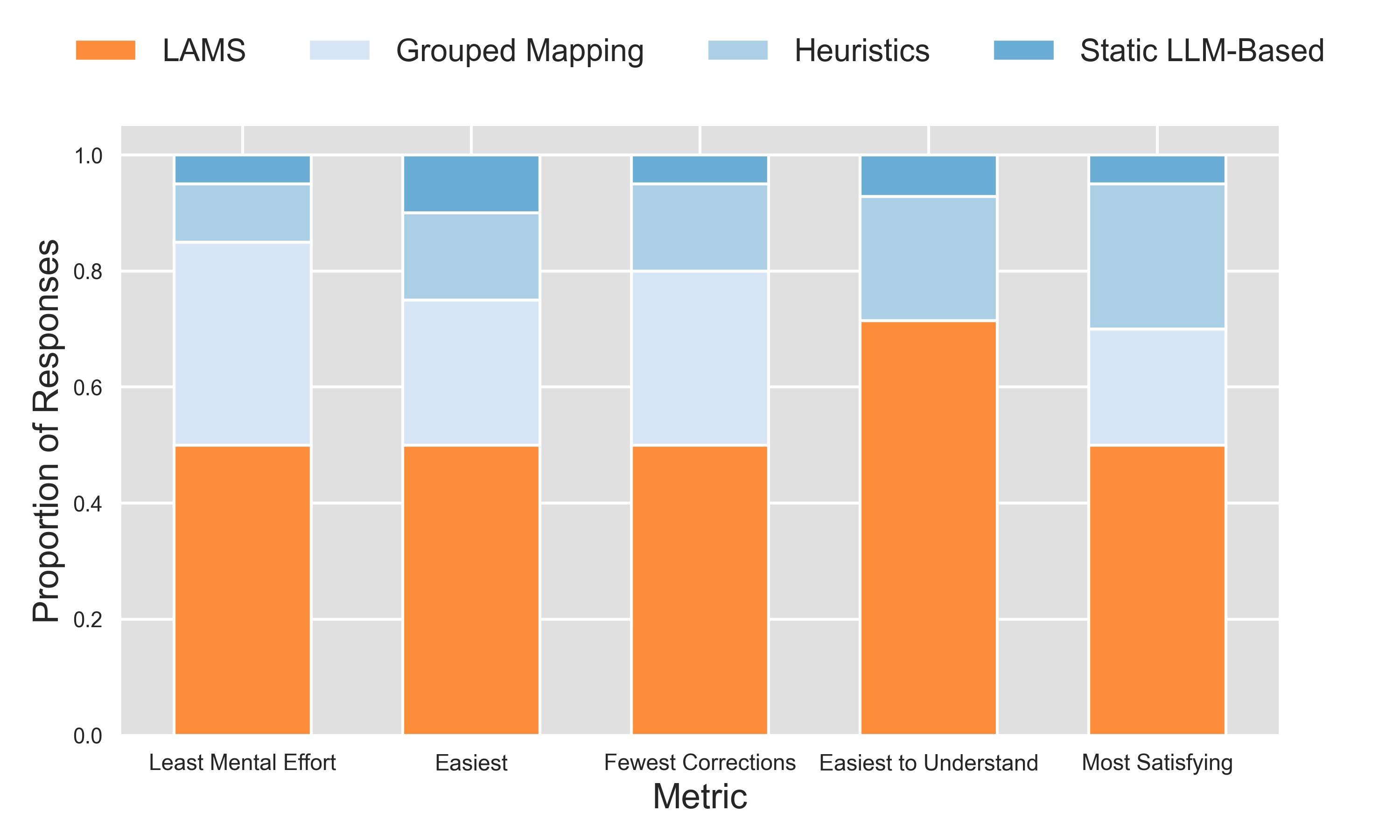} 
\caption{Proportion of participants who preferred each method in trial 3, based on the combined responses from both tasks.}
\label{fig:preferences}
\vspace{-1.5em}
\end{figure}

\subsubsection{\textbf{User Preferences}}
In addition to recording the number of manual mode switches, we asked participants to report their preferences for the four methods after the third trial. The order of mode-switching methods was counterbalanced in each trial. As Grouped Mapping requires a distinct manual mode-switching approach, users were informed when using this method, but for the other three methods, the specific method being used was not disclosed to users. Participants answered questions regarding (1) which method they felt required the least mental effort, (2) was easiest to complete the task with, (3) required the fewest corrections, (4) was easiest to understand when and why mode switching occurred, and (5) was the most satisfying method to use. Note that given the multi-trial, multi-method nature of our study, administering a detailed questionnaire like the NASA TLX after each trial would greatly increase participant burden. To minimize this burden and ensure data quality, we opted for the approach of collecting preference data only once per task after trial 3.

The proportion of participants who preferred each method for these metrics, based on the combined responses from both tasks in trial 3, is shown in Fig. \ref{fig:preferences}. 
For the fourth question, we excluded responses that preferred Grouped Mapping, where all mode switching are manual, making this question inapplicable.

As observed in Fig.~\ref{fig:preferences}, LAMS was preferred by more participants across all five questions compared to the other methods. Multinomial tests with a null hypothesis of equal user preference across all methods showed significant deviation for all five metrics ($p < 0.001$), supporting the statement in H1 that LAMS is preferred over alternative methods. Further discussions on the reasons behind user preferences, as revealed by our post-study interviews, are provided in Appendix \ref{sec:interview}.

\section{Discussion}
\label{sec:discussion}

Our experimental results demonstrate the effectiveness of the LAMS framework. As a preliminary step towards utilizing LLMs for automatic mode switching, in this section, we examine the strengths and limitations of LAMS, focusing on when LLM-based switching works well and when it encounters difficulties. We also discuss potential future directions to address these challenges and further improve the system.

\subsubsection*{\textbf{LAMS’s Advantage over a Static Method: Impact of Incremental Improvement in the First Trial}}
As shown in our user study, LAMS consistently outperforms the static LLM-based mode-switching method.

Notably, even in the first trial, LAMS shows improvements during the later stages of the tasks by making more accurate mapping predictions and requiring fewer manual mode switches. Specifically, in the first trial of the water pouring task, the static LLM-based method required an average of 8.6 manual mode switches, whereas LAMS reduced this to 7.5, representing a 12.8\% improvement. Similarly, in the book storage task, the static method required 9.1 manual switches, while LAMS reduced this to 6.3, a 30.8\% improvement.

While the incremental improvement of LAMS over the static LLM-based method was expected after multiple trials, the improved performance observed in the first trial was an unexpected benefit. To understand this, we examined the errors made by both methods during the tasks.

We found that LAMS's advantage largely stems from reducing the number of incorrect joystick-to-robot action mappings for the ``Open Gripper'' and ``Close Gripper'' commands. The static LLM-based method frequently mapped the joystick's up movement to ``Open Gripper'' or the down movement to ``Close Gripper'' when these actions were not relevant, whereas LAMS made far fewer such errors.

Specifically, in the first trial of the water pouring task, the static method made an average of 6.7 incorrect ``Open Gripper'' or ``Close Gripper'' mappings, while LAMS reduced this to 3.7. In the book storage task, the static method averaged 5.2 false mappings, while LAMS lowered this to just 1.8. False ``Open Gripper'' or ``Close Gripper'' mappings are defined as instances where the joystick's up or down movements were mapped to gripper actions, but the user manually switched to a different robot action, indicating the error.

\subsubsection*{\textbf{Challenges in Differentiating Rotational Movements}}

While LAMS effectively reduces manual mode switches, we found that it struggled to differentiate between certain rotational movements during the user study.

Specifically, LAMS often confused ``Yaw Left/Right'' and ``Roll Left/Right'' (both corresponding to lateral joystick movements), even by the third trial when the system had more opportunities to learn from user-generated examples. In contrast, because the longitudinal joystick movements correspond only to one type of rotation, ``Pitch Up/Down'', LAMS performed better in predicting this motion.

In particular, LAMS mapped joystick movements to ``Pitch Up/Down'' 80\% of the time when the action was required. However, it achieved 40\% accuracy for ``Yaw Left/Right'' and 50\% for ``Roll Left/Right''. Correct predictions were defined where the joystick's movements were mapped to the corresponding rotational action, and the user executed the action, confirming its correctness. The total number of times when a rotational action was required includes both correct predictions and cases where the joystick's movements were mapped to a different action but were manually switched to the corresponding rotational movement by the user.

This challenge likely arises from the inherent complexity of representing and interpreting 3D rotations in both natural language and mathematical contexts~\cite{hinckley1997usability, hong20233d}. Future work could focus on exploring alternative ways to express 3D rotations in natural language to mitigate this challenge. Further discussion on the limitations of LAMS can be found in Appendix \ref{sec:limitations}.

\subsubsection*{\textbf{Conclusion}}
In this paper, we present LAMS, which leverages LLMs to perform automatic mode switching for teleoperating a robotic arm. LAMS is able to incrementally improve as a user repeatedly interacts with the system. Our ablation and user studies demonstrate that LAMS outperforms alternative designs, baseline mode-switching methods, and a static LLM-based approach. It enables users to complete tasks with fewer manual mode switches, is preferred by users, and improves performance with use over time.

\bibliographystyle{IEEEtran}
\balance
\bibliography{hri}

\clearpage
\begin{appendices}
\renewcommand{\thesubsection}{\thesection.\arabic{subsection}}

\section{Grouped Mapping: Details and Explanation}
\label{sec:grouped}
One of the baseline methods we tested in our user study, \textbf{Grouped Mapping}, is a common mode-switching method in robotic arm applications \cite{harmonic, losey2022learning, latent0, yang2023high, padmanabha2023hat, padmanabha2024independence, gopinath2017mode, herlant2016assistive}, where robot actions are divided into predefined groups. For example, in group 1, the joystick controls move forward/backward and move left/right. In group 2, it controls move up/down and roll left/right. Group 3 maps to pitch up/down and yaw left/right, while group 4 maps the longitudinal movements to gripper open/close, with no lateral mapping. Users switch between these groups. Note that unlike other methods we tested in our user study, Grouped Mapping uses a different manual mode-switching mechanism to preserve group integrity: as shown in Fig.~\ref{fig:interface}, the user presses the X button on the Xbox controller to cycle through groups, changing all four joystick mappings at once. This single-button cycling method was used in to reflect the most commonly used mode-switching approach in current teleoperation systems \cite{harmonic, losey2022learning, latent0, yang2023high, padmanabha2023hat, padmanabha2024independence, gopinath2017mode, herlant2016assistive}.

\section{Experimental Setup \& Graphical User Interface (GUI)}
\label{sec:gui}
\subsection*{B.1 Experimental Setup}
Figure B1 shows the experimental setup of the user study. As shown in the figure, the participant is seated behind the robotic arm. Throughout the study, the user can continuously see a GUI displaying the current control mode, which shows the four robot action directions mapped to each joystick movement.

\begin{figure}[h]
    \centering
    \includegraphics[width=0.8\linewidth]{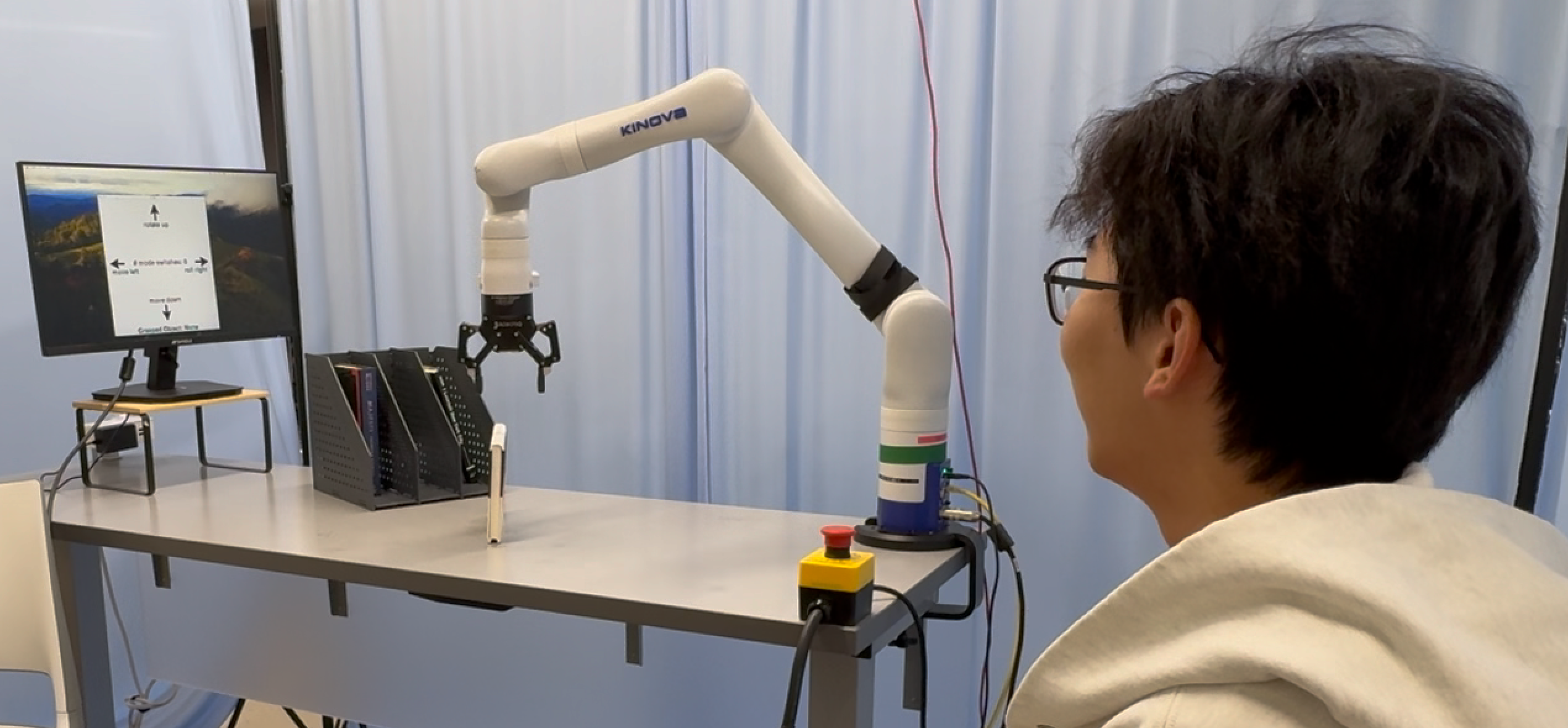}
    \caption*{Figure B1: Experimental setup of our user study.}
\end{figure}

\subsection*{B.2 Graphical User Interface (GUI)}
An example of the Graphical User Interface (GUI) displayed to participants during the experiments is shown in Figure B2.1. As shown in the figure, the GUI indicates the four robot action directions currently mapped to each joystick movement. It also displays the number of manual mode switches performed by the participant and the object currently being grasped, allowing the experimenters to more easily supervise and record the progress of the experiments.

An example of how the GUI updates after an automatic mode switch initiated by the LLM is shown in Figure B2.2. The LLM predicts the most likely robot actions mapped to the four joystick directions. If the predicted robot action for a particular joystick movement differs from the action assigned before the LLM call, the corresponding text on the GUI is highlighted in blue. This visual cue is designed to alert the user that the robot action associated with that direction has changed.
\begin{figure}[h]
    \centering
    \includegraphics[width=0.4\linewidth]{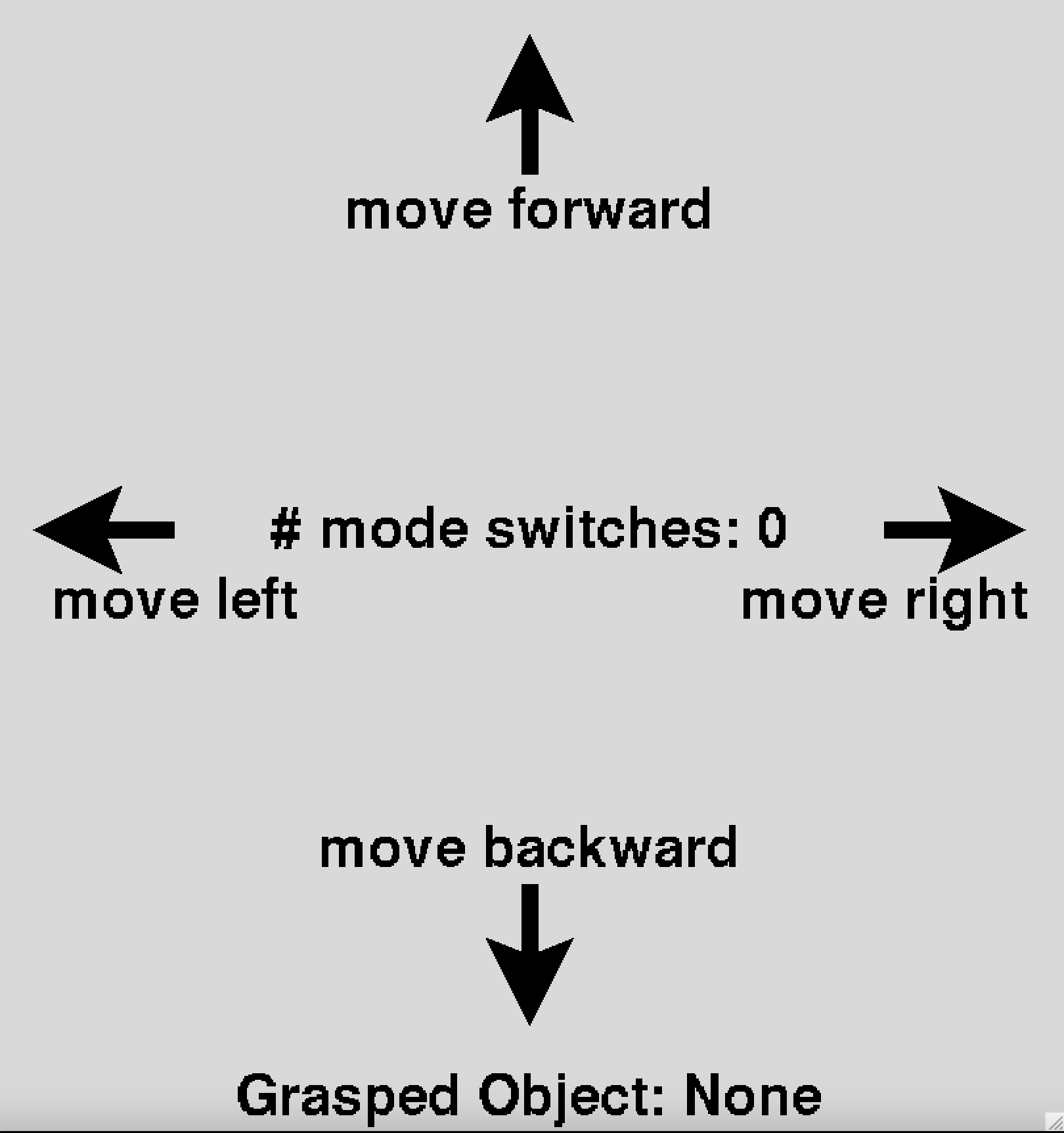}
    \caption*{Figure B2.1: An example of the Graphical User Interface (GUI) displayed to participants during the experiments.}
\end{figure}

\begin{figure}[h]
    \centering
    \includegraphics[width=0.4\linewidth]{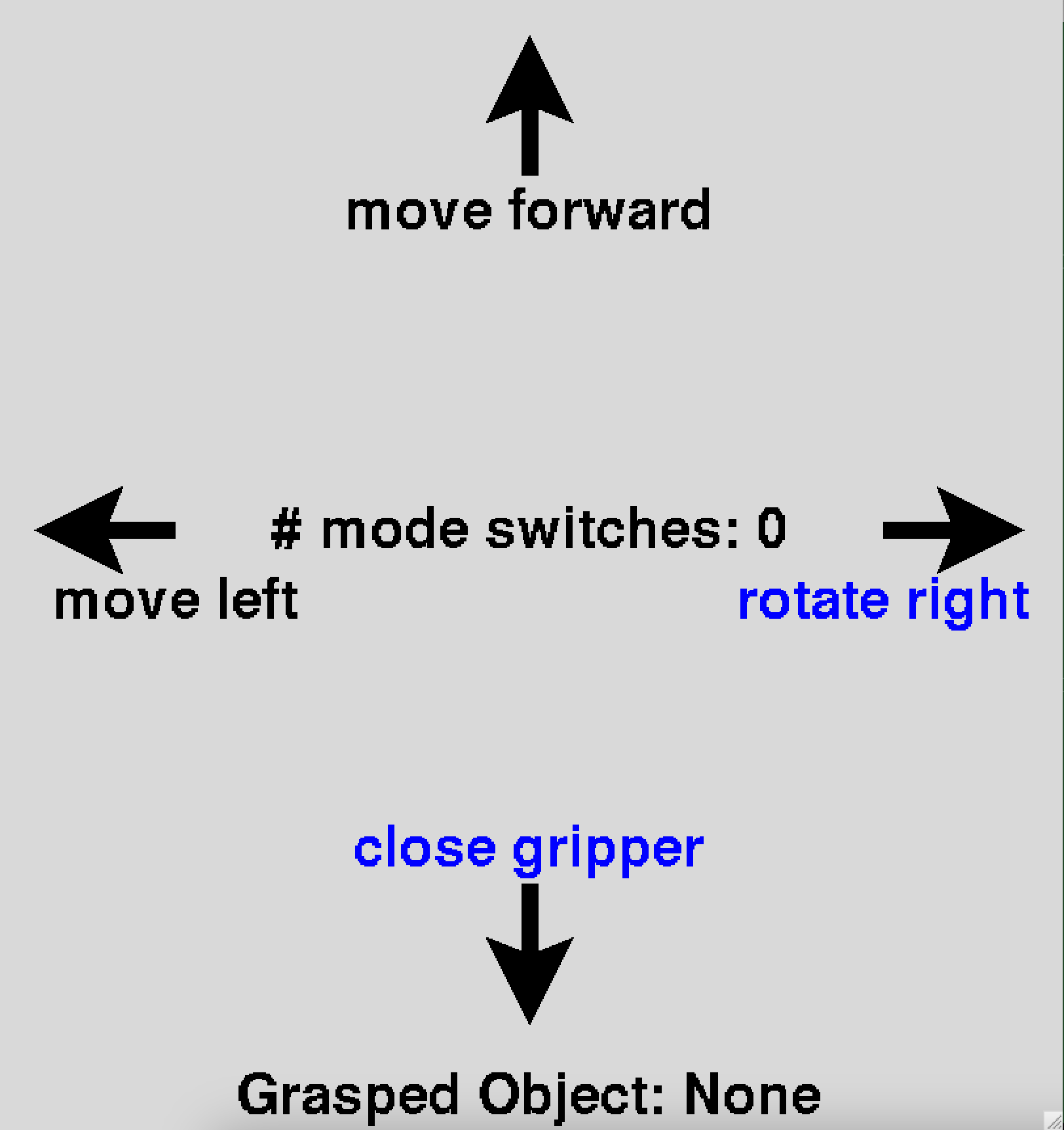}
    \caption*{Figure B2.2: An example of how the GUI updates after an automatic mode switch initiated by the LLM.}
\end{figure}

An example of how the GUI updates after a manual mode switch in experiments using LAMS, Static LLM-Based Mode Switching, and Hand-Engineered Heuristic Switching is shown in Figure B2.3. In these three methods, when the user is dissatisfied with the current action mappings, they can press one of the directional buttons on the Xbox controller's D-pad. This updates the mapping for the corresponding joystick direction without affecting the others. The updated mapping is highlighted in red on the GUI.

\begin{figure}[h]
    \centering
    \includegraphics[width=0.4\linewidth]{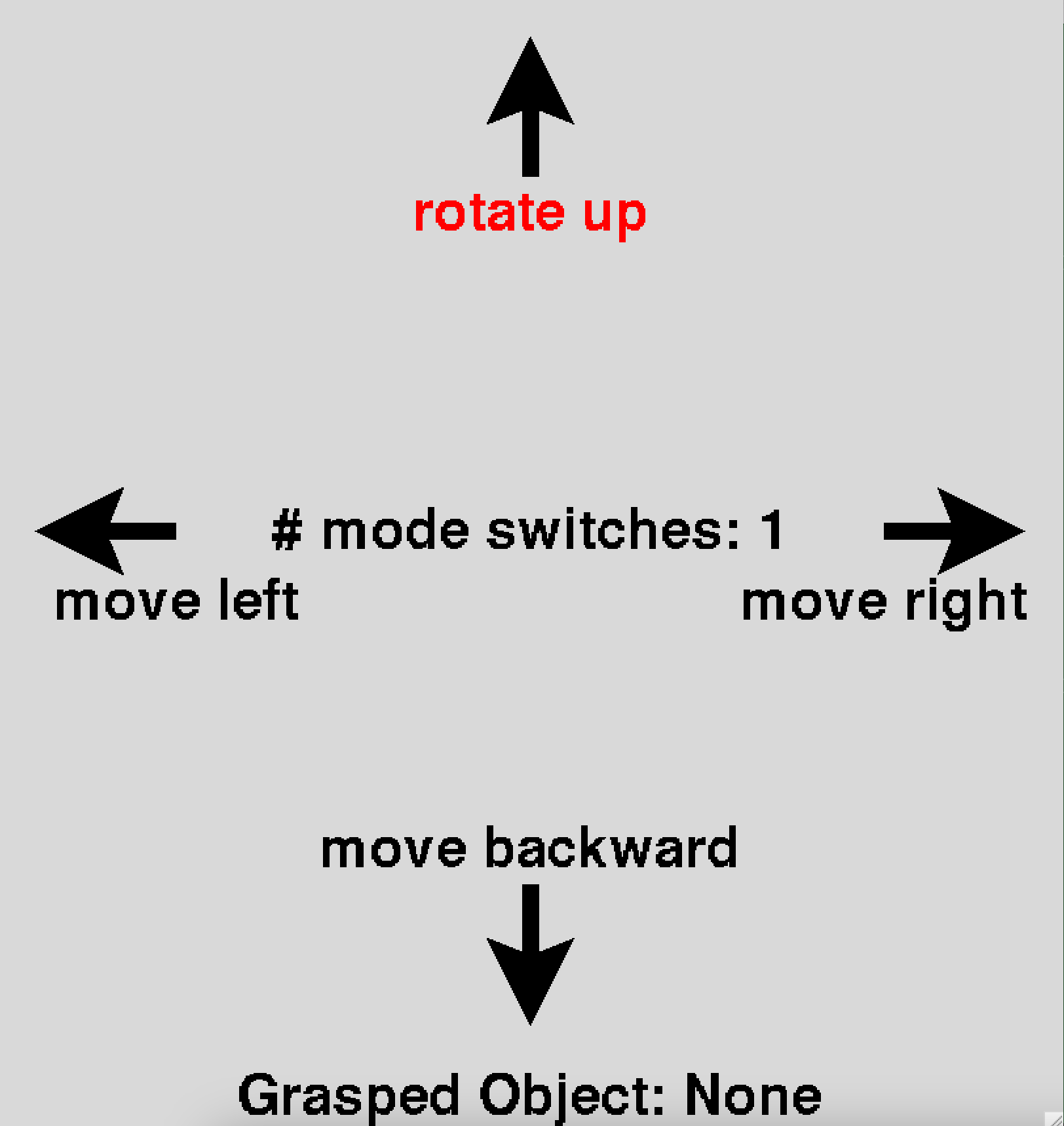}
    \caption*{Figure B2.3: An example of how the GUI updates after a manual mode switch in experiments using LAMS, Static LLM-Based Mode Switching, and Hand-Engineered Heuristic Switching.}
\end{figure}

An example of how the GUI updates after a manual mode switch in experiments using Grouped Mapping is shown in Figure B2.4. To maintain group integrity, the user presses the X button on the Xbox controller to cycle through predefined groups, updating all four joystick mappings simultaneously. In this case, the number of manual mode switches increases by only one, regardless of the number of joystick mappings changed.

\begin{figure}[h]
    \centering
    \includegraphics[width=0.4\linewidth]{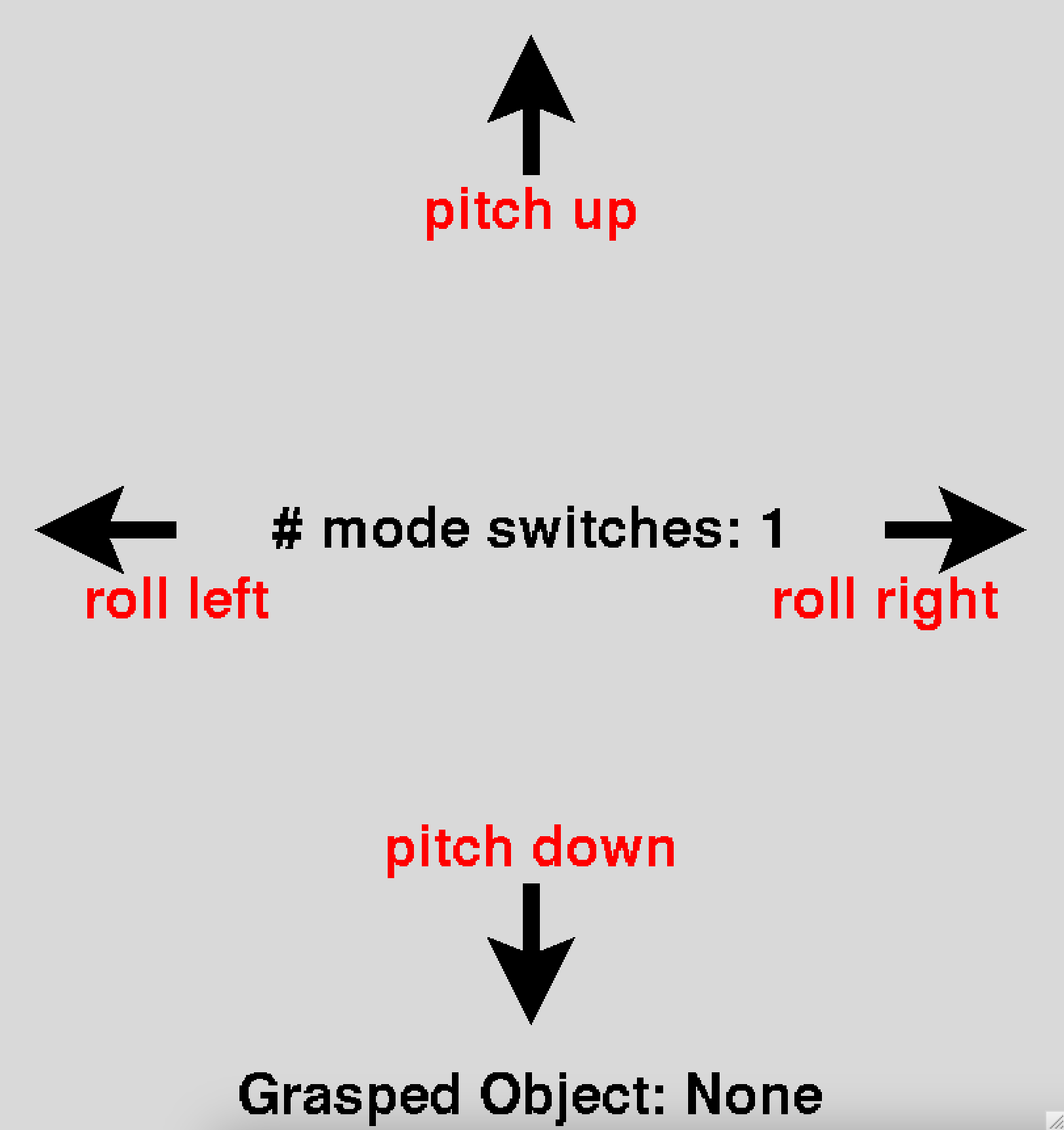}
    \caption*{Figure B2.4: An example of how the GUI updates after a manual mode switch in experiments using Grouped Mapping.}
\end{figure}

\begin{figure*}[t]
    \centering
    \begin{subfigure}[b]{0.4\textwidth} 
        \centering
        \includegraphics[width=\linewidth]{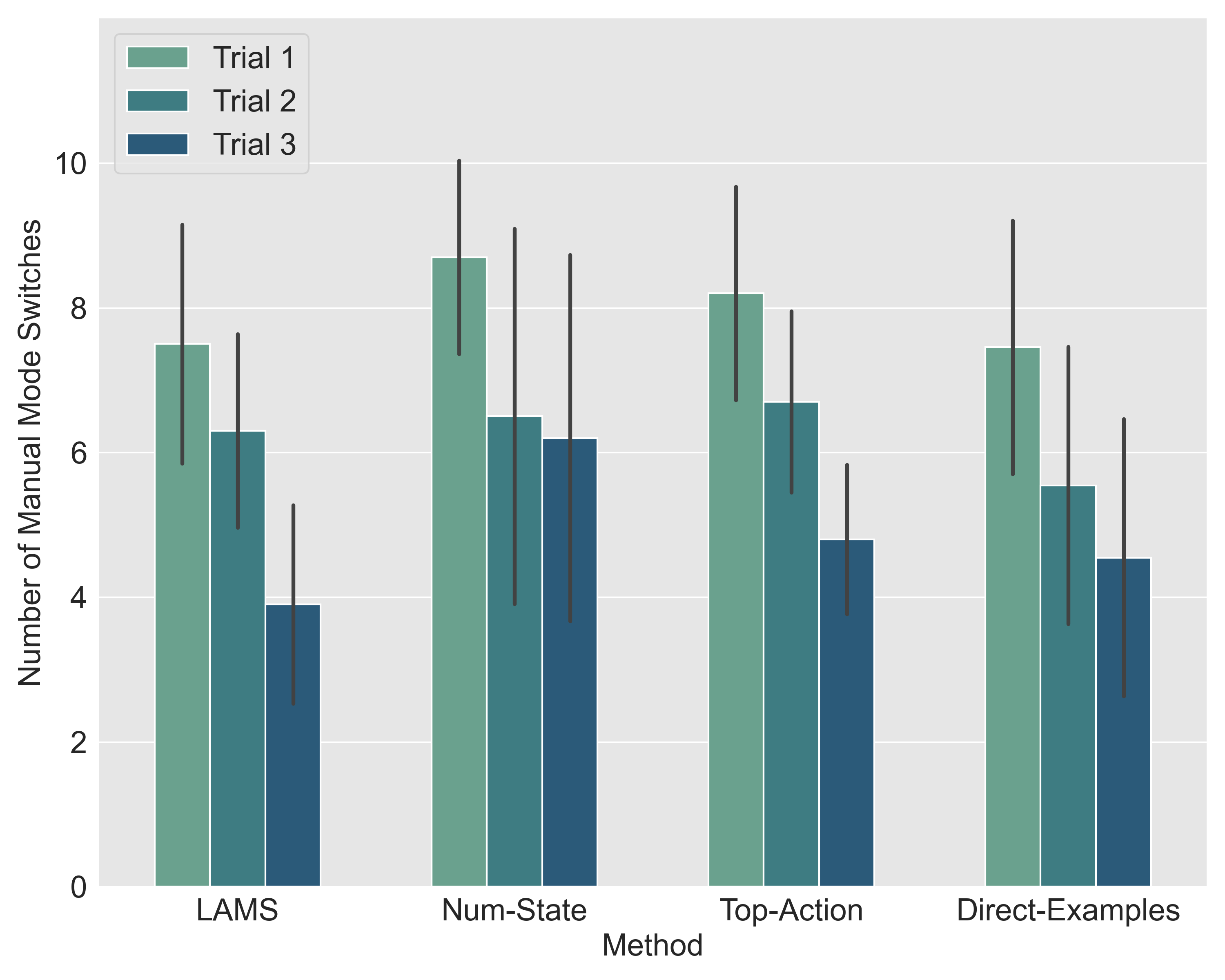}
        \caption{Water Pouring task}
        \label{fig:mode-switches-jar}
    \end{subfigure}
    \hspace{0.05\textwidth}
    \begin{subfigure}[b]{0.4\textwidth} 
        \centering
        \includegraphics[width=\linewidth]{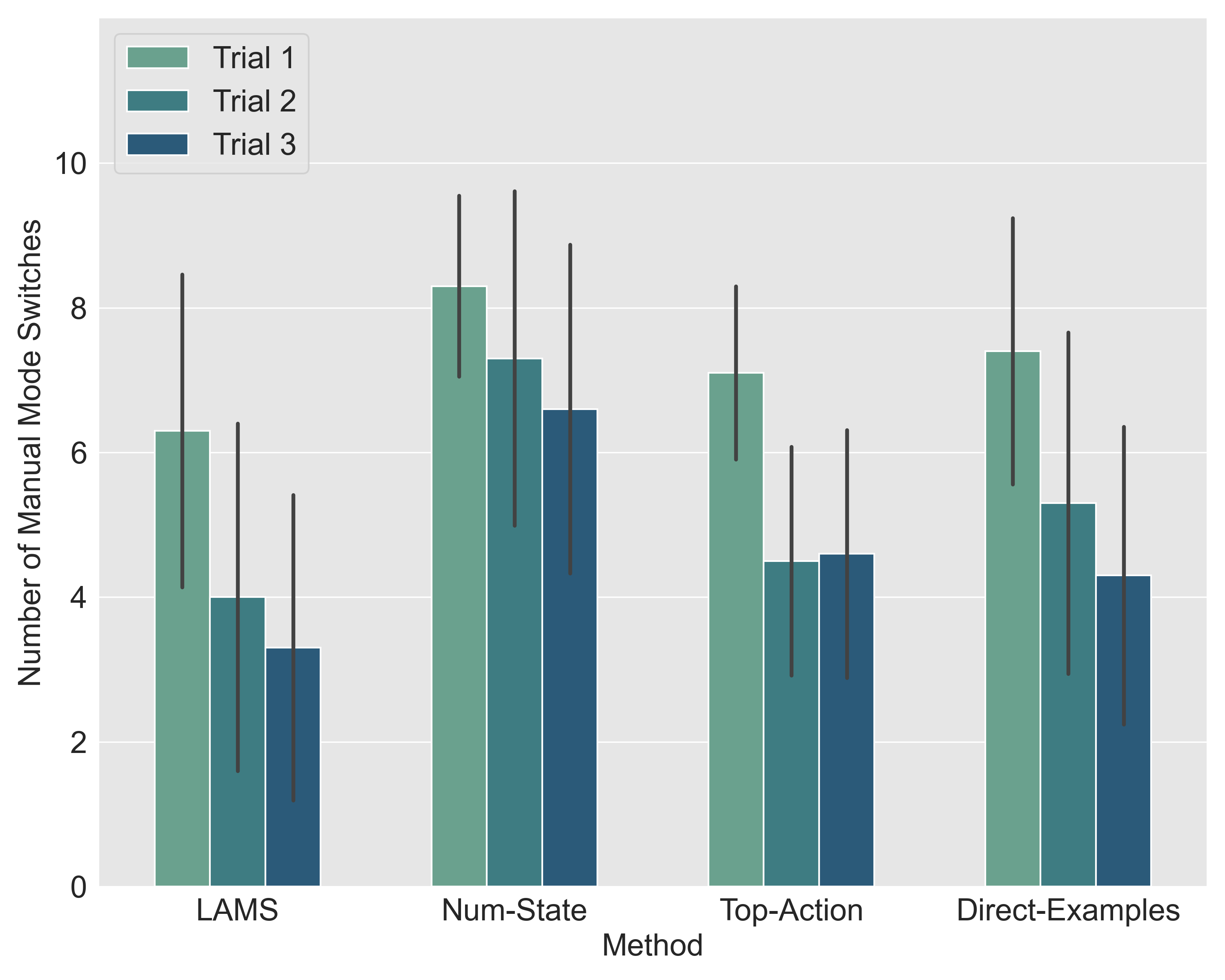}
        \caption{Book Storage task}
        \label{fig:mode-switches-book}
    \end{subfigure}
    \caption*{Figure C: Simulated number of manual mode switches calculated through ``shadow mode'' analysis, averaged over 10 participants in the user study. Error bars show standard deviations.}

\end{figure*}

\section{Ablation Study Extension: ``Shadow Mode'' Analysis on User Study Data}
\label{sec:shadow}

To further validate the effectiveness of the key design choices in LAMS described in Section \ref{sec:ablation}, we conducted a ``shadow mode’’ analysis based on data collected from our user study. Specifically, at each instance where LAMS invoked the LLM to predict a control mode, we made parallel LLM calls with prompts corresponding to the respective ablated methods. For each simulated method, if the user’s subsequent action deviated from the predicted control mapping, we assumed that a manual mode switch would have been required if the participant had been using that method. These simulated corrections were also considered as examples to be integrated into subsequent LLM calls for incremental improvement of the corresponding methods. The average number of manual mode switches calculated through this approach across the 10 participants in the user study is presented in Figure C.

It is important to note that this "shadow mode" analysis provides an approximation rather than a precise evaluation of the ablated methods. Participants’ behavior and decision-making processes may vary when actively using different methods, potentially leading to differences in task execution routes and mode-switching patterns. Nevertheless, this analysis offers meaningful insights by evaluating the relative efficiency of each method under the same task scenarios and interaction points.

As shown in Figure C, consistent with the findings in Section \ref{sec:ablation}, LAMS achieves the lowest average number of manual mode switches among all methods, with a low standard deviation. This result reaffirms the effectiveness of the key design choices in LAMS.

\section{Reasons Behind User Preferences Revealed by Our Post-Study Interview}
\label{sec:interview}

Following our user study, we conducted oral interviews to explore participants' reasons for their preferences regarding different mode-switching methods.

As shown in Fig. \ref{fig:preferences}, LAMS was preferred by more participants across all five questions compared to the other methods. In post-study interviews, participants noted that they favored LAMS for its responsiveness and its ability to accurately switch modes when needed, especially in comparison to the Heuristic method, which only switched modes at predefined subtask transitions. Particularly, for our fourth question concerning which method was easiest to understand, participants mentioned that their confusion primarily stemmed from incorrect mode-switching predictions. Since LAMS provided the most accurate predictions, participants found it the easiest to understand. 

Nonetheless, some participants preferred Grouped Mapping over other mode-switching methods for its full manual control without algorithmic assistance. In post-study interviews, these participants noted that repeated task performance in a short period allowed them to memorize an optimal mode-switching sequence, making full manual control more intuitive. However, in real-world scenarios with varied or intermittent tasks, memorizing optimal sequences may be less feasible, and algorithmic assistance could be more advantageous. 

These findings suggest that while reducing manual switches enhances efficiency, user preferences may be influenced by task familiarity and control. Future work could explore these aspects to better understand user experiences.

\section{Limitations of Our Work}
\label{sec:limitations}
One potential limitation of our work is our text-based task context grounding approach, which may face challenges in more complex environments or when object relevance is ambiguous. Future work could test the method’s adaptability in environments with multiple, ambiguous objects.

We also acknowledge the potentially higher cost of LLM calls compared to methods like maintaining example libraries and applying nearest-neighbor searches. Compared to alternative approaches, our LLM-driven method likely provides greater generalizability, especially with limited examples and dynamic contexts. However, this comparison was not formally conducted in our work. Future studies could explore these comparisons and examine the trade-offs between cost and the generalizability of LLMs.

\section{Prompts Used in Our Method}
\label{sec:prompts}
\subsection*{F.1 Prompt Prefix $l_{pre}$}

\begin{lstlisting}
**Objective:**  
You will be given task instructions, the current state of the robot arm, and information of objects around. Your goal is to predict the most likely actions out of the specified groups of actions.  

**Data Structures:**
1. **Current State of the Robot Arm:**
- **Type:** Dictionary
- **Keys:**
    - `position`: A dictionary indicating the coordinates of the robot arm's position in centimeters.
        - `x`: The position along the x-axis, an integer value in centimeters.
        - `y`: The position along the y-axis, an integer value in centimeters.
        - `z`: The position along the z-axis, an integer value in centimeters.
    - `orientation`: A dictionary indicating the orientation of the robot arm in degrees.
        - `roll`: The rotation around the x-axis, an integer value in degrees ranging from 0 to 360.
        - `pitch`: The rotation around the y-axis, an integer value in degrees ranging from 0 to 360.
        - `yaw`: The rotation around the z-axis, an integer value in degrees ranging from 0 to 360.
    - `gripper`: A string `open` or `closed` indicating whether the gripper is open or closed.

2. **Object Information:**
- **Type:** Dictionary
- **Keys:** The object type as a string.
- **Values:** 
    - A dictionary containing:
        - `relative_pos`: Either a natural language string "The robot arm is holding the object." or "has been dropped", or a dictionary with two keys `relative_position` and `relative_orientation`. 

            For `relative_position`, the dictionary should have three keys `x_relation`, `y_relation`, and `z_relation`, each containing a natural language string describing the object's position relative to the robot arm in the respective direction. For example:
            
            - `x_relation`: "to the forward of the robot arm" or "to the backward of the robot arm" or "close to the robot arm along the x-axis"
            - `y_relation`: "to the left of the robot arm" or "to the right of the robot arm" or "close to the robot arm along the y-axis"
            - `z_relation`: "above the robot arm" or "below the robot arm" or "close to the robot arm along the z-axis"
            
            For `relative_orientation`, the dictionary should have three keys `pitch_relation`, `roll_relation`, and `yaw_relation`, each containing a natural language string describing the object's orientation relative to the robot arm in the respective axis. For example:

            - `pitch_relation`: "pitched more up compared to the robot arm" or "pitched more down compared to the robot arm" or "pitch orientation is close to the robot arm's pitch orientation"
            - `roll_relation`: "rolled more left compared to the robot arm" or "rolled more right compared to the robot arm" or "roll orientation is close to the robot arm's roll orientation"
            - `yaw_relation`: "yawed more left compared to the robot arm" or "yawed more right compared to the robot arm" or "yaw orientation is close to the robot arm's roll orientation"
    

  


**Task:**

Based on the provided information and the current task, robot state, and object information, determine the most likely actions from each of the following groups. For each group, the actions are labeled with identifiers A, B, and C for clarity. Groups 1 and 2 also include an additional action labeled as D.
---

**Definition of Most Likely Actions:**  
Most likely actions refer to the actions that have the highest probability of successfully achieving the task objectives based on the current state of the robot arm, information of objects around, and the specified action groups. These actions should be determined by evaluating the robot's ability to manipulate objects effectively and efficiently according to the given criteria.

---

**Output Requirements:**  
Your output should be a dictionary where each key represents a group, and the corresponding value is the most likely action's letter identifier (A, B, C, or D for groups 1 and 2, and A, B, or C for groups 3 and 4) followed by the corresponding action description. The output should look like this, do not output any additional analysis:

{
"Group 1": "A/B/C/D: {corresponding most likely action from group 1}",
"Group 2": "A/B/C/D: {corresponding most likely action from group 2}",
"Group 3": "A/B/C: {corresponding most likely action from group 3}",
"Group 4": "A/B/C: {corresponding most likely action from group 4}",
}

**Group 1:**
- A: Move forward
- B: Move up
- C: Rotate up
- D: Open gripper
        
**Group 2:**
- A: Move backward
- B: Move down
- C: Rotate down
- D: Close gripper
        
**Group 3:**
- A: Move left
- B: Roll left
- C: Rotate left

**Group 4:**
- A: Move right
- B: Roll right
- C: Rotate right

---
\end{lstlisting}

\subsection*{F.2 An instance of Rule Prompt $l_{rule}^{t}$}

\begin{lstlisting}
Below are a set of rules derived from previous examples. These rules summarize the patterns identified between task information, robot arm's state, object information, and the chosen actions. Your task is to apply these rules to predict the most likely actions out of the specified groups for the current situation.

1. **Gripper State and Object Proximity:**
   - If the robot arm's gripper is open and an object (e.g., a bottle cap) is close to the robot arm along all axes (x, y, z), the most likely action is to adjust the orientation to align with the object's orientation. This may involve actions like "Pitch up" if the object is pitched more up compared to the robot arm.

2. **Object Proximity and Gripper Action:**
   - When an object, such as a bottle cap, is "close to the robot arm" in all axes and the gripper is open, the most likely action is to close the gripper to grasp the object. This is typically the first step in manipulating an object that is within reach.

3. **Rule for Opening the Gripper:**
   - If the robot arm is holding the bottle cap and the gripper is closed, the most likely action is to open the gripper. This action is necessary to release the cap, which is a prerequisite for picking up the bottle.

4. **Relative Positioning for Pouring:**
   - When the task involves pouring contents from one object (e.g., a bottle) into another (e.g., a bowl), the robot should ensure the pouring object is positioned above the receiving object. The robot should adjust its orientation to facilitate the pouring action, ensuring the gripper is closed to maintain a secure hold on the pouring object.

5. **General Positioning and Orientation Considerations:**
   - The robot arm's actions are influenced by the relative position and orientation of the objects. If an object is close to the robot arm, the arm should adjust its orientation to align with the object's orientation for effective manipulation.
   - The gripper state (open or closed) is crucial in determining the next action. If the gripper is closed and holding an object, the next action is likely to release it. Conversely, if the gripper is open, the next action is likely to involve positioning the arm to grasp another object.

6. **Rule for Group 1 Actions:**
   - If the robot arm's gripper is open and the bottle cap is close to the robot arm in all axes (x, y, z), and the cap's pitch is more up compared to the robot arm, the most likely action is to "Pitch up" to align the robot arm's orientation with the cap for effective manipulation.

7. **Roll Adjustments for Object Orientation:**
   - When an object is "rolled more right compared to the robot arm", the robot may need to "roll right" to align with the object's orientation. Conversely, if the object is "rolled more left", the robot may need to "roll left".

8. **Horizontal Position Adjustments:**
   - If an object like a bowl is "to the forward of the robot arm" and the task involves moving towards it, the robot may need to "move forward". If the object is "to the backward of the robot arm", the robot may need to "move backward".

9. **Vertical Position Adjustments:**
   - If an object like a bottle is "below the robot arm" and the task involves interacting with it, the robot may need to "move down" to align with the object. Conversely, if the object is "above the robot arm", the robot may need to "move up".

10. **Task Sequence Consideration:**
   - The sequence of actions is determined by the task requirements. For tasks involving multiple steps, such as opening a cap and then pouring, the robot arm must first complete the initial step (e.g., opening the cap) before proceeding to the next (e.g., picking up the bottle). The actions are chosen to ensure the task progresses logically and efficiently.

11. **Rule for Group 1 Actions:**
   - If the robot arm is holding the bottle cap (gripper is closed), and the cap is not in the robot's immediate vicinity, the most likely action is to move the robot arm upwards. This is to lift the cap away from the bottle, indicating the completion of the cap removal task.

12. **Task-Specific Actions:**
   - For tasks that involve pouring, such as pouring contents from a bottle into a bowl, the robot arm may need to "Roll left" or "Roll right" to achieve the correct pouring angle, especially if the bottle is already being held.

13. **Relative Position and Task Execution:**
   - If an object (e.g., a bottle) is close to the robot arm along the x and y axes but below it along the z-axis, the robot should prepare to pick up the object by adjusting its orientation to match the object's roll and yaw. This ensures a secure grip and effective manipulation.

14. **Rule for Group 2 Actions:**
   - If the robot arm's gripper is open, the bottle cap has been dropped, and the bottle is close to the robot arm along the x and y axes but below it along the z-axis, the most likely action is to "Pitch down" to align the robot arm's orientation with the bottle for picking it up.

16. **Rule for Group 3 Actions:**
   - **Condition:** When the robot arm is holding the bottle (indicating the gripper is closed), and the bowl is positioned close to the robot arm along the x-axis, to the left along the y-axis, and below along the z-axis, with the pitch orientation closely aligned and the roll orientation slightly different (e.g., rolled more right), the most likely action is to roll the arm left. This action aligns the bottle for pouring into the bowl.

16. **Rule for Moving Forward:**
   - If the task involves interacting with an object that is positioned "close to the robot arm along the x-axis" and "close to the robot arm along the y-axis," and the object is "below the robot arm" or "close to the robot arm along the z-axis," the most likely action is to "Move forward." This is particularly applicable when the object is directly in front of the robot arm and the gripper is open, indicating readiness to engage with the object.
   
These rules are designed to guide the robot arm in effectively interacting with the bottle and its cap, ensuring that the tasks of opening the cap, picking up the bottle, and pouring its contents are completed efficiently.

\end{lstlisting}

\subsection*{F.3 Details on the Construction of Robot End Effector and Task Objects Pose Description $l_{pose}^{t}$ and an Instance}

As described in Section \ref{sec:input}, $l_{pose}^{t}$ provides a description of the current pose of the robot arm and task-relevant objects. For the robot arm, we encode its pose as a dictionary containing the end-effector's Cartesian coordinates and Euler angles, along with the gripper status. Cartesian coordinates are expressed in centimeters and Euler angles in degrees. To enhance interpretability and improve the effectiveness of the LLM, we discretize the continuous values into integer intervals—5 cm for Cartesian coordinates and 15 degrees for Euler angles. This discretization method is inspired by prior works~\cite{discretize, rt2}, which have shown discretization to enhance the effectiveness of LLM-based systems. The discretization granularity was chosen empirically for robust performance. Finer levels were empirically less stable.

The gripper is represented simply with natural languages as either ``open'' or ``closed''.

For task objects, we describe the relative position of each object with respect to the robot arm's end-effector across six dimensions, using six natural language statements: $l_{obj_{i},x}^{t}$, $l_{obj_{i},y}^{t}$, $l_{obj_{i},z}^{t}$, $l_{obj_{i},\text{roll}}^{t}$, $l_{obj_{i},\text{pitch}}^{t}$, and $l_{obj_{i},\text{yaw}}^{t}$. For example, $l_{obj_{i},y}^{t}$ might be ``to the left of the robot arm'' or ``close to the robot arm'' where ``close'' is defined as within 5 cm for Cartesian coordinates and within 15 degrees for Euler angles, consistent with our discretization scheme. If the object is ``close'' to the end effector in all six dimensions, these six statements are simplified to a single statement: ``The robot arm is holding the {object}''. Through ablation studies (Section~\ref{sec:ablation}) we find that this natural language grounding of object states is more effective than using numeric representations in LAMS. 

Below is an instance of $l_{pose}^{t}$ from our experiments:

\begin{lstlisting}
### Current Task, Robot Arm State, and Object Information:   

- **Current Task:** Open the cap of a bottle, then pick up the bottle and pour what's inside into a bowl.

- **Current State of the Robot Arm:**  
{
    "position": {
        "x": 40,         
        "y": 35,
        "z": 20
    },
    "orientation": {
        "theta x": 180,
        "theta y": 0,
        "theta z": 90
    }
    "gripper": open
}

- **Current Object Information:**  
{
    "bottle cap": {
        "relative_pos":"The robot arm is holding the bottle cap.",    
    },
    "bottle": {
        "relative_pos":{
            "relative_position":{
                "x_relation": "to the forward of the robot arm",
                "y_relation": "to the right of the robot arm",
                "z_relation": "below the robot arm",
            },
            "relative_orientation":{
                "pitch_relation": "pitched more down compared to the robot arm",
                "roll_relation": "roll orientation is close to the robot arm's roll orientation",
                "yaw_relation": "yaw orientation is close to the robot arm's roll orientation",
            },
        }
    },
    "bowl": {
        "relative_pos":{
            "relative_position":{
                "x_relation": "to the forward of the robot arm",
                "y_relation": "to the right of the robot arm",
                "z_relation": "below the robot arm",
            },
            "relative_orientation":{
                "pitch_relation": "pitched more down compared to the robot arm",
                "roll_relation": "roll orientation is close to the robot arm's roll orientation",
                "yaw_relation": "yaw orientation is close to the robot arm's roll orientation",
            },
        }
    },
}

- **Output (do not output any additional analysis):**  
{
"Group 1": "A/B/C/D: {corresponding most likely action from group 1}",
"Group 2": "A/B/C/D: {corresponding most likely action from group 2}",
"Group 3": "A/B/C: {corresponding most likely action from group 3}",
"Group 4": "A/B/C: {corresponding most likely action from group 4}",
}
\end{lstlisting}

\subsection*{F.4 An instance of Mode-Switching Example $l_{e}^{t}$ Generated from User-Interaction}
\begin{lstlisting}
**Example 0:**   

- **Current Task:** Open the cap of a bottle, then pick up the bottle and pour what's inside into a bowl.

- **Current State of the Robot Arm:**  
{
    "position": {
        "x": 50,         
        "y": 30,
        "z": 15
    },
    "orientation": {
        "theta x": 120,
        "theta y": 0,
        "theta z": 90
    }
    "gripper": open
}

- **Current Object Information:**  
{
    "bottle cap": {
        "relative_pos":{
            "relative_position":{
                "x_relation": "close to the robot arm along the x-axis",
                "y_relation": "close to the robot arm along the y-axis",
                "z_relation": "close to the robot arm along the z-axis",
            },
            "relative_orientation":{
                "pitch_relation": "pitched more up compared to the robot arm",
                "roll_relation": "rolled more right compared to the robot arm",
                "yaw_relation": "yaw orientation is close to the robot arm's roll orientation",
            },
        }
    },
    "bottle": {
        "relative_pos":{
            "relative_position":{
                "x_relation": "close to the robot arm along the x-axis",
                "y_relation": "close to the robot arm along the y-axis",
                "z_relation": "below the robot arm",
            },
            "relative_orientation":{
                "pitch_relation": "pitched more down compared to the robot arm",
                "roll_relation": "rolled more right compared to the robot arm",
                "yaw_relation": "yaw orientation is close to the robot arm's roll orientation",
            },
        }
    },
    "bowl": {
        "relative_pos":{
            "relative_position":{
                "x_relation": "to the forward of the robot arm",
                "y_relation": "to the left of the robot arm",
                "z_relation": "below the robot arm",
            },
            "relative_orientation":{
                "pitch_relation": "pitched more down compared to the robot arm",
                "roll_relation": "rolled more right compared to the robot arm",
                "yaw_relation": "yaw orientation is close to the robot arm's roll orientation",
            },
        }
    },
}

- **Most Likely Action(s):**  
{
"Group 1": "C: Pitch up"
}
\end{lstlisting}

\subsection*{F.5 Prompt Prefix for Rule Generation $l_{pre-rule}$}
\begin{lstlisting}
**Objective:**  
You will be given examples of task instructions, poses of a robot arm, and information of objects around it. 
Your goal is to analyze the examples and summarize the patterns or rules, which will be used to assist another agent to predict the most likely actions out of the specified groups of actions in similar scenarios of the same task.    


**Data Structures:**
1. **Current State of the Robot Arm:**
- **Type:** Dictionary
- **Keys:**
        - `position`: A dictionary indicating the coordinates of the robot arm's position in centimeters.
            - `x`: The position along the x-axis, an integer value in centimeters.
            - `y`: The position along the y-axis, an integer value in centimeters.
            - `z`: The position along the z-axis, an integer value in centimeters.
        - `orientation`: A dictionary indicating the orientation of the robot arm in degrees.
            - `theta_x`: The rotation around the x-axis, an integer value in degrees ranging from 0 to 360.
            - `theta_y`: The rotation around the y-axis, an integer value in degrees ranging from 0 to 360.
            - `theta_z`: The rotation around the z-axis, an integer value in degrees ranging from 0 to 360.
    - `gripper`: A string `open` or `closed` indicating whether the gripper is open or closed.

2. **Object Information:**
- **Type:** Dictionary
- **Keys:** The object type as a string.
- **Values:** 
    - A dictionary containing:
        - `relative_pos`: Either a natural language string "The robot arm is holding the object." or "has been dropped", or a dictionary with two keys `relative_position` and `relative_orientation`. 

            For `relative_position`, the dictionary should have three keys `x_relation`, `y_relation`, and `z_relation`, each containing a natural language string describing the object's position relative to the robot arm in the respective direction. For example:
            
            - `x_relation`: "to the forward of the robot arm" or "to the backward of the robot arm" or "close to the robot arm along the x-axis"
            - `y_relation`: "to the left of the robot arm" or "to the right of the robot arm" or "close to the robot arm along the y-axis"
            - `z_relation`: "above the robot arm" or "below the robot arm" or "close to the robot arm along the z-axis"
            
            For `relative_orientation`, the dictionary should have three keys `pitch_relation`, `roll_relation`, and `yaw_relation`, each containing a natural language string describing the object's orientation relative to the robot arm in the respective axis. For example:

            - `pitch_relation`: "pitched more up compared to the robot arm" or "pitched more down compared to the robot arm" or "pitch orientation is close to the robot arm's pitch orientation"
            - `roll_relation`: "rolled more left compared to the robot arm" or "rolled more right compared to the robot arm" or "roll orientation is close to the robot arm's roll orientation"
            - `yaw_relation`: "yawed more left compared to the robot arm" or "yawed more right compared to the robot arm" or "yaw orientation is close to the robot arm's roll orientation"
    

The task of the agent you are trying to help is to determine the most likely actions from each of the following groups, based on the provided current robot state and object information:
---

**Definition of Most Likely Actions:**  
Most likely actions refer to the actions that have the highest probability of successfully achieving the task objectives based on the current state of the robot arm, information of objects around, and the specified action groups. These actions should be determined by evaluating the robot's ability to manipulate objects effectively and efficiently according to the given criteria.

---

### Examples:
\end{lstlisting}

\end{appendices}
\end{document}